%% file: main.tex
\documentclass[11pt]{article}

\usepackage{ACL2023}

\usepackage{times}
\usepackage{latexsym}

\usepackage[T1]{fontenc}

\usepackage[utf8]{inputenc}

\usepackage{microtype}

\usepackage{inconsolata}

\usepackage{url}            
\usepackage{booktabs}       
\usepackage{amsfonts}       
\usepackage{nicefrac}       
\usepackage{xcolor}         
\usepackage{amsmath}
\usepackage{graphicx}
\usepackage{comment}
\usepackage{enumitem}
\usepackage{amsmath}
\usepackage{float} 
\usepackage{algorithm}
\usepackage[noend]{algorithmic}
\usepackage{xcolor} 
\newcommand{\TODO}[1]{$ $\newline\noindent\colorbox{yellow!30}{\parbox{\dimexpr\the\columnwidth-2\fboxsep}{\textbf{\texttt{TODO:}} \textit{#1}}}}
\newcommand{\STORY}[1]{$ $\newline\noindent\colorbox{blue!30}{\parbox{\dimexpr\the\columnwidth-2\fboxsep}{\textit{#1}}}}

\usepackage{xcolor, soul}

\usepackage{placeins} 

\usepackage{array}
\usepackage{multirow}
\usepackage{hhline}

\author{
Michael Lan\textsuperscript{\dag}\textsuperscript{*} \quad Philip Torr\textsuperscript{\ddag} \quad Fazl Barez\textsuperscript{\dag \ddag}\textsuperscript{*} \\
\textsuperscript{\dag} Apart Research \\
\textsuperscript{\ddag} Department of Engineering Sciences, University of Oxford
}

\begin{document}

\definecolor{darkgreen}{rgb}{0.0, 0.6, 0.0}
\definecolor{gold}{rgb}{0.83, 0.69, 0.22}
\definecolor{lightgray}{gray}{0.65}

\title{Towards Interpretable Sequence Continuation: \\Analyzing Shared Circuits in Large Language Models}
\maketitle
\begin{abstract}
While transformer models exhibit strong capabilities on linguistic tasks, their complex architectures make them difficult to interpret. Recent work has aimed to reverse engineer transformer models into human-readable representations called circuits that implement algorithmic functions. We extend this research by analyzing and comparing circuits for similar sequence continuation tasks, which include increasing sequences of \emph{Arabic numerals}, \emph{number words}, and \emph{months}. By applying circuit interpretability analysis, we identify a key sub-circuit in both GPT-2 Small and Llama-2-7B responsible for detecting sequence members and for predicting the next member in a sequence. Our analysis reveals that semantically related sequences rely on shared circuit subgraphs with analogous roles. Additionally, we show that this sub-circuit has effects on various math-related prompts, such as on intervaled circuits, Spanish number word and months continuation, and natural language word problems. 
This mechanistic understanding of transformers is a critical step towards building more \textit{robust}, \textit{aligned}, and \textit{interpretable} language models. 


\end{abstract}

\def\thefootnote{*}\footnotetext{ Equal contribution}\def\thefootnote{\arabic{footnote}}

\input{texs/1_intro}

\input{texs/3_related} 
\input{texs/4_methods} 
\input{texs/5_seq_conn}
\input{texs/6_seq_func}
\input{texs/7_llama2_seqcont}

\input{texs/8_math_behavior}
\input{texs/conclusion}

\bibliography{bibliography}
\bibliographystyle{acl_natbib}

\appendix
\input{texs/apdx_expm_setup}
\input{texs/apdx_gpt_results}
\input{texs/apdx_llama_results}

\input{texs/apdx_abl_gen}
\input{texs/appendix}

\end{document}

%% file: texs/1_intro.tex
\section{Introduction}
\label{sec:intro}
Transformer-based large language models (LLMs) like GPT-4 have demonstrated impressive natural language capabilities across a variety of tasks \citep{brown2020language,bubeck2023sparks}. 
However, these models largely remain black boxes due to their complex, densely connected architectures. Understanding how these models work is important for ensuring safe and aligned deployment, especially as they are already being used in high-impact real-world settings \citep{Zhang2022ShiftingML,caldarini2022literature,micelibarone2023larger}.

Several researchers argue that the ability to interpret AI decisions is essential for the safe implementation of sophisticated machine learning technologies \citep{hendrycks2022x,barez2023iii}. Previous studies show that AI interpretability is vital for AI safety, for catching deception, and for addressing misalignment \citep{BarredoArrieta2020ExplainableAI,Amodei2016ConcretePI}. 
Mechanistic interpretability, a sub-field of interpretability, aims to reverse engineer models into understandable components (such as neurons or attention heads) \citep{elhage2021mathematical}. By uncovering underlying mechanisms, researchers can better predict model behaviors \citep{mu2020compositional, foote2023neuron} and understand emergent phenomena \citep{nanda2023progress,quirke2023understanding, marks2023interpreting}. 

To analyze computations within  models, a recent approach has been to find \emph{circuits}, which are subgraphs of neural networks that represent algorithmic tasks \citep{elhage2021mathematical}.
Recent work in interpretability has uncovered transformer circuits that implement simple linguistic tasks, such as identifying indirect objects in sentences \citep{wang2022interpretability}. However, only a few studies have focused on the existence of shared circuits \citep{merullo2023circuit}, in which circuits utilize the same sub-circuits for similar tasks. 
Identifying shared circuits assists in aligning AI via methods such as model editing \citep{meng2023locating}, which precisely targets problematic areas for more efficient re-alignment without erroneously altering healthy components. Documenting the existence of shared circuits enables safer, more predictable model editing with fewer risks, as editing a circuit may affect another if they share sub-circuits \citep{hoelscherobermaier2023detecting}. Therefore, interpretability enables safer alignment by understanding adverse effect prevention.


While models use the same components for different tasks, such as when there are far more tasks/features than neurons \citep{elhage2022superposition}, our focus is on locating components which are shared due to similar, re-usable functionality, and not for vastly different functionalities. Our work tackles the hypothesis that LLMs may re-use circuits across analogous tasks that share common abstractions. For instance, 
similar sequence continuation tasks, such as number words ("one two three") and months ("Jan Feb Mar"), can be analogously mapped to one another via the natural number abstraction (for example, one and Jan are mapped to 1). As these tasks share a common abstraction, LLMs may have learned to efficiently re-use components that utilize shared patterns. Understanding how LLMs re-use components based on commonalities can shed light on how they represent and associate semantic concepts with one another \citep{gurnee2023language}. Not only would this enhance understanding of how LLMs actually perceive information, but it may have potential applications in transfer learning \citep{zhuang2020comprehensive}.

Thus, in this paper, we demonstrate the existence of shared circuits for similar sequence continuation tasks, as the  similarity across these tasks is clear, allowing us to cleanly pinpoint functionality. Our key finding is that there exist shared sub-circuits between similar tasks in GPT-2 Small \citep{radford_language_2019} and Llama-2-7B \citep{touvron2023llama2openfoundation}, where the shared components have the same functionality across tasks. 

The main contributions of this work are\footnote{To encourage reuse and further development our code and datasets can be found here: \url{https://github.com/apartresearch/seqcont\_circuits}}:

\begin{enumerate}[label=\textbf{\arabic*})]
    \item \textbf{The discovery of shared circuits for similar sequence continuation tasks.} 
    \item \textbf{The finding of similar sub-circuits across models with analogous functionality.}
    \item \textbf{Showing that these circuits affect natural language math-related prompts.}
    
\end{enumerate}


\begin{figure*}[ht]
  \centering
  \includegraphics[scale=0.2]{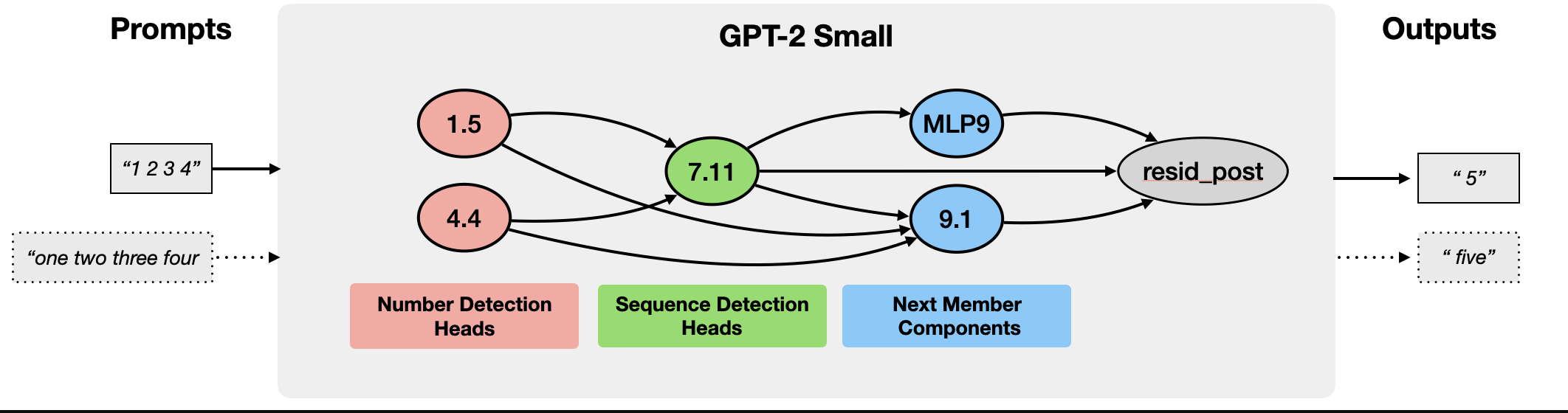}  
  \caption{
  The important components of a shared, entangled sub-circuit for the Numerals, Number Words, and Months tasks in GPT-2 Small. The functional roles of the components are labeled below them. Resid\_post denotes the residual stream state right before the linear unembedding to logits. Full circuits are shown in Appendix \ref{appendix:full_circs}. }
  \label{fig:all_no_qkv_circ}
\end{figure*}

%% file: texs/3_related.tex
\section{Background and Related Work}


\textbf{Transformer Models.} 
A \textit{transformer model} learns the importance of different parts of the data in relation to each other via query-key attention mechanisms.
We analyze LLM transformer-based models that take in text inputs as token sequences $(x_1, \ldots, x_L)$ of length $L$.
Tokens are mapped to $d$-dimensional embeddings 
by a function $\Psi : \mathcal{D} \rightarrow \mathbb{R}^d$
\citep{vaswani2017attention}.


\emph{Attention Head.} A transformer consists of blocks of attention heads, each consisting of two matrices: the query-key $\mathbf{\textbf{QK}}$ matrix that outputs the attention pattern $A_{i,j}$, and the output-value $\mathbf{\textbf{OV}}$ matrix that outputs to the residual stream. An attention layer output is the sum of attention heads $h_{i,j}$. We use the notation L.H for attention heads, where L is a layer index and H is a head index in layer L.

\emph{Multi-Layer Perceptron.} Each attention layer output is passed to a Multi-Layer Perceptron (MLP). The MLPs in transformers are generally made of two linear layers, parameterized by weight matrices $W_1$ and $W_2$, with a ReLU activation function in between. 




\emph{Residual Stream.} Attention head and MLP outputs are added to the residual stream, from which components read from and write to. Components in non-adjacent layers are able to interact via indirect effects from the additivity of the residual stream \citep{elhage2021mathematical}.

\textbf{Transformer Circuit Discovery.} 
In transformer circuits, evidence has shown that in general, MLPs associate input information with features \cite{geva2020transformer}, while attention heads move information \citep{olsson2022context}.

Prior work has employed \textbf{causal interventions} to locate circuits for specific tasks \citep{meng2023locating, vig2020investigating}, such as for the Indirect Object Identification (IOI) task, in which the goal is to complete sentences with the correct subject \citep{wang2022interpretability}. One type of causal intervention is called \textbf{knockout}, which, after a model has processed a dataset, replaces (or \emph{ablates}) the activations of certain components with other values, such as activations sampled from another distribution. The sampled activations may come from a \emph{corrupted dataset}, which outputs the wrong answer, but resembles the same dataset without the information of interest (for example "1 2 3" becomes "8 1 4" to preserve information about numbers, while removing sequence information). After running again, if the ablated nodes do not change model performance much, they are deemed as not part of a circuit of interest.

Another type of causal intervention is \textbf{activation patching}, which takes the corrupted dataset as input, and then restores the activations at a certain component with the original activations to observe how much that restored component recovers the original performance. 
The Automatic Circuit DisCovery (ACDC) technique employs iterative patching automatically find circuit graphs for given tasks \citep{conmy2023towards}; however, this technique only seeks to automate finding the connectivity of circuit graphs, and not their functionality interpretation.

\vspace{5mm}

\textbf{Interpretability of Sequential Tasks.}
In concurrent work, successor heads were found by \cite{gould2023successor} to increment tokens with a natural ordering. 
However, \cite{gould2023successor} only analyze the output of the successor head, rather than looking at the output of a model, and does not look for circuits for sequences with more than one member. As shown in our paper, a successor head is not enough to evaluate a sequence. Additionally, we study shared interactions between components. Thus, we build on \cite{gould2023successor} by showing how successor heads interact with the model as a whole.
\cite{hanna2023does} found circuits for "greater-than" sequence tasks (for example completing the sentence, “The war lasted from the year 1732 to the year 17”, with any valid two-digit end years > 32). Greater-than tasks allow any year greater than a value to be valid, which differs from our sequence completion tasks that only have one valid answer. The authors noted that "similar tasks had similar... circuits", but compared across number tasks, and not across non-number tasks such as months. In our work, we study tasks that are more dissimilar.
Related to sequence continuation, \cite{stolfo-etal-2023-mechanistic} and \citep{quirke2023understanding} discover circuits for arithmetic computation.

\textbf{Shared Circuits for Similar Tasks.} Locating shared circuits is a relatively new research topic
. Previous studies have noted that circuits for the Induction task \citep{olsson2022context} are found in circuits for the IOI task. Recently, \cite{merullo2023circuit} discovered shared circuits for the IOI task and Colored Objects task (where the aim is to identify the correct color for an object given a set of colored objects). The authors utilized an intervention experiment to improve the Colored Objects circuit by modifying subject inhibition heads of the IOI circuits to inhibit the wrong color answers. In our paper, we focus on tasks which are much more similar and map to a common abstraction. While the IOI task and Colored Objects task both share similar sub-tasks such as "inhibiting tokens", the focus of our paper is on enhancing our understanding of how LLMs represent analogous concepts by discovering sub-circuits which represent common abstractions, instead of just shared sub-tasks.


%% file: texs/4_methods.tex
\section{Methodology}
\label{sec:methods}

\textbf{Circuit Discovery Process. } Our approach begins by applying iterative pruning to obtain connectivities for circuits of similar tasks. Then, we employ methods to deduce component functionalities shared by similar tasks. We approach circuit discovery in two types of stages:
\footnote{The methods we apply to one stage may also yield information about another stage. }


\begin{enumerate}
    \item \emph{Connectivity Discovery} consists of applying causal mediation analysis techniques for identifying important connections for varying component granualarity levels (eg. residual stream, attention head, MLP, neuron). 
    \item \emph{Functionality Discovery} aims to describe the tasks handled by circuit components, labeling them with interpretable semantics. 
\end{enumerate}



\subsection{Connectivity Discovery Methods}
\label{sec:connMethods}

\textbf{Metrics. } We utilize the \emph{logit difference} to measure model task capability by taking the difference between the correct token $L_{C}$ and an incorrect token logit $L_{I}$. The incorrect logit may be chosen as a token that is not the correct token.
To compare an ablated model with the unablated model, we employ the \textbf{performance score}, a percentage calculated as the logit difference of the ablated model over the logit difference of the unablated model. 
We define an \textit{important component} to be a component that, when ablated, greatly reduces performance.






\textbf{Iterative Pruning for Nodes.} To search for circuit components, we use a knockout method that ablates one candidate component, or node, $n_c$ at a time and checks how much performance falls. This method begins with all the components as a set $C$, which are nodes of a \emph{candidate circuit}. Let $N$ denote the set of all the nodes in the network. At each step, ablation is performed by patching in the mean activations of a corrupted dataset at $N \setminus C$, which are the nodes that are not part of of the current candidate circuit. Then, the mean activations of the corrupted dataset are also patched in at new candidate node $n_c$. If performance falls below $T_n$, a user-defined \emph{performance threshold}, node $n_c$ is kept in the candidate circuit $C$, as it is deemed necessary for the task. Else, it is removed.

We start by selecting component $n_c$ from the last layer, continuing until the first layer; we call this procedure the \emph{backward sweep}. At each layer during the backward sweep, we first ablate the layer's MLP, and then consider its attention heads. Next, we then prune again from the first layer to the last layer; we call this the \emph{forward sweep}. At each layer during the forward sweep, we first ablate each attention heads, and then its MLP. We continue iterating by successive backward-forward sweeps, stopping when no new components are pruned during a sweep. The output is $C$, the unpruned node set.

This method may be considered as a simplified and coarser variation of ACDC \citep{conmy2023towards}, which decomposes heads into key, query, and value (qkv) vector interactions.
As head outputs deemed unimportant may also be unimportant when decomposed, our method first filters nodes at a coarse level, then decomposes heads into separate (qkv) nodes during edge pruning.
\footnote{While the graphs found by ACDC utilize even finer granularity levels than just head decomposition, the authors of the paper note that different granularity levels are valid based on analysis goals (eg. \cite{hanna2023does} analyze at a level without head decomposition). We find our chosen granularity level to be sufficient for analyzing shared circuits.}

\textbf{Iterative Path Patching for Edges.} We describe this part of the method in Appendix \ref{edge_patching}.

\subsection{Functionality Discovery Methods}
\label{funcmethods}



\textbf{Attention Pattern Analysis. } We analyze the $\mathbf{QK}$ matrix of attention heads to track information movement from keys to queries. We take the mean of datasets samples to calculate the attention scores of the heatmaps in this paper, and display only one sample on the axes for demonstration purposes.


\textbf{Component Output Scores.} We analyze head outputs by examining the values written to the residual stream via the heads' output matrices ($\mathbf{OV}$), allowing us to see what information is being passed by each head along in the circuit. These values are measured by component output scores; we utilize a \textbf{successor score} that measures how well a head, given sequence token $I$, outputs token $I+1$. The details of this method are described in Appendix \ref{appendix:fmethod}.

\textbf{Logit Lens.} Logit lens is a method for understanding the internal representations by unembedding each layer output into vocabulary space and analyzing the top tokens \citep{nostalgebraist2020interpreting}. We use logit lens to uncover the layer at which the predicted token goes from the 'last sequence member' to the 'next sequence member'.

%% file: texs/5_seq_conn.tex
\section{Circuit Connectivity in GPT-2 Small}
\label{sec:connExpms}

\subsection{Experimental Setup}


We test on GPT-2 Small, which has 144 heads and 12 MLPs, with a total of 117M parameters.

\textbf{Task Comparison.} We compare increasing sequences of: (1) Arabic Numerals (or 'Numerals'), (2) Number Words, and (3) Months. 

\textbf{Datasets.} We run a generated prompts dataset of length 4 sequences  (eg. 1 2 3 4). Our focus in this paper is not on finding circuits only for Numeral sequences, but on prompt types that share a common abstraction. Thus, to better compare numbers to months, we use sequences ranging from 1 to 12. 

For each task on GPT-2 Small, we generate samples by placing our sequence members among non-sequence tokens. For instance, one sample may be 'Kyle was born in February. Bob was born in March. Grant was born in April. Madison was born in'. Placing sequence members amongst non-sequence tokens allows us to evaluate the circuit representation of the shared sub-task of how the model selects sequence members from non-sequence members. We generate a total of 1536 samples per task, with a total of 4608 samples.
More details can be found in Appendix \ref{appendix:dataset}. 

\emph{Corrupted Datasets.} We corrupt sequence information by using randomly chosen tokens of a similar sequence type (eg. `1 2 3' is replaced with `8 1 4'). The non-sequence tokens are kept the same, while the sequence members are replaced. 

\textbf{Metric. } We measure using logit difference using the last sequence member as the incorrect token (eg. for "1 2 3", the correct token is "4", the incorrect token is "3").

\subsection{Shared Sub-Circuits for Similar Sequence Continuation Tasks}


In this section we analyze main results, and include extended results in Appendix \S \ref{appendix:gpt2conn}.
We discover shared sub-circuits across the three sequence continuation tasks. These circuits were found using a performance threshold of $T_n = T_e = 80$\%. Figure \ref{fig:all_no_qkv_circ} shows a sub-circuit found across the circuits for all three tasks, which includes attention heads 4.4, 7.11 and 9.1, which we show to be important in Table \ref{circHeadsDrop}. Figure \ref{fig:full_all_no_qkv_circ} in Appendix \ref{appendix:full_circs} combines all three circuits into one graph  \footnote{Due to the (qkv) circuit's large display size, we show the circuits with (qkv) decomposition in Appendix \ref{appendix:indiv_circ}.}.

Several important attention heads are identified across various circuits. We define an attention head as \textbf{important} if their ablation from a task's circuit causes an average performance drop of at least 20\% for all tasks \footnote{20\% is chosen due to using $T_n=80\%$, so that for many removal order variations, a component with a 20\% importance cannot be removed, unless there are alternative backups.}. Table \ref{circHeadsDrop} compares the importance of these attention heads for our tasks. We note that ablating attention heads 0.1, 4.4, 7.11, and 9.1 cause drop of more than 20\% for all three circuits, while ablating attention head 1.5 causes a drop of approximately 20\% for Numerals and Number Words circuits.


\begin{table*}
    \centering
    \caption{Drop in Task Performance for GPT-2 Small when an important attention head is removed from the circuit in Figure \ref{fig:full_all_no_qkv_circ}.}
    \label{circHeadsDrop}
    \vspace{0.5em}
    \begin{tabular}{c|ccc}
        \toprule
        Important Head & Numerals & NumWords & Months \\
        \midrule
        0.1 & -44.29\% & -78.74\% & -52.10\% \\
        4.4 & -33.19\% & -34.11\% & -73.16\% \\
        7.11 & -41.64\% & -44.78\% & -45.37\% \\
        9.1 & -34.94\% & -27.74\% & -43.03\% \\
        1.5 & -27.83\% & -18.65\% & - \\
        \bottomrule
    \end{tabular}
\end{table*}


\textbf{MLP Connectivity. }
For all tasks, we find that several MLP ablations cause a >20\% performance drop. In particular, MLP 9 causes a substantial drop of more than 90\%. These results are found in Appendix \ref{appendix:MLP}.



%% file: texs/6_seq_func.tex
\section{Explaining Shared Component Functionalities in GPT-2 Small}
\label{sec:connFunc}

In this section we analyze main results, and include extended results in Appendix \S \ref{appendix:gpt2func}.

\subsection{Sub-Circuit Hypothesis}
\label{subcirc_hypo}
We hypothesize the important shared components for the three tasks work together as a \emph{functional} sub-circuit. We define sub-tasks that all three tasks share: 
(1) Identifying Sequence Members and (2) Predicting the Next Member.

Our hypothesis is that early attention heads, in particular 1.5 and 4.4, identify similar, adjacent sequence members, such as numbers or Months, without yet attending to the distinction of which numbers should be focused on more than others. Following this, information is passed further along the model to other components such as attention head 7.11, which may discern that the two most recent elements are more significant. This information is then conveyed to attention head 9.1 to put more emphasis on predicting the next element in the sequence. Lastly, the next element calculation is done primarily by MLP 9. Thus, this sub-circuit would represent an algorithm that carries out the sub-tasks shared for all three tasks.
This section details evidence that supports this circuit hypothesis.


\subsection{Sequence Member Detection Heads}
\label{sec:earlyHeads}

\begin{figure*}[ht]  
  \centering
 \includegraphics[scale=0.175]{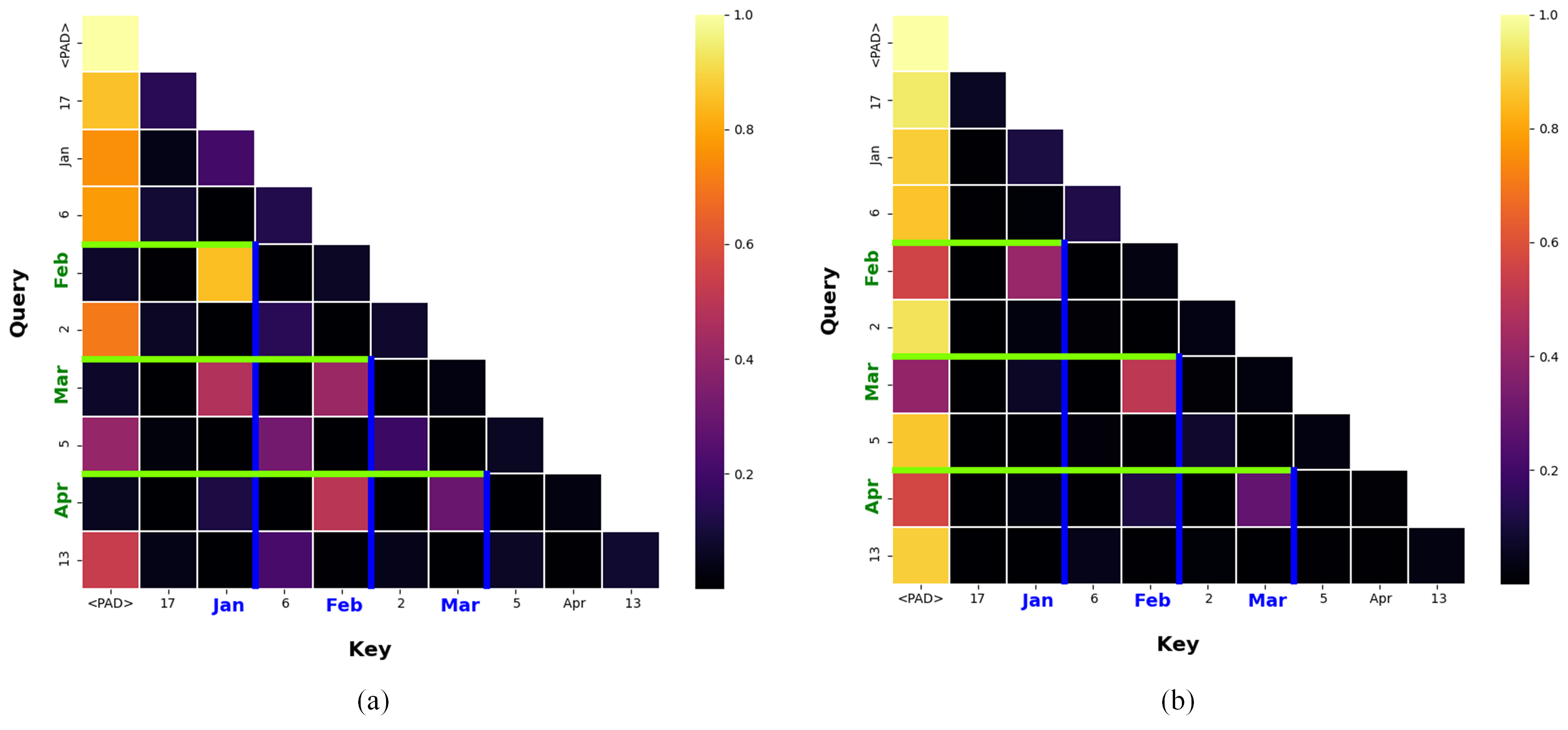}
  \caption{GPT-2 Small attention patterns for (a) Attention Head 1.5 and (b) Head 4.4. Lighter colors mean higher attention values. For each of these detection patterns, the \textcolor{darkgreen}{query is shown in green}, and the \textcolor{blue}{key is shown in blue}. The Months are in sequential order, but the Numerals are not. For attention head 1.5, similar types attend to similar types. But for head 4.4, Months attend to Months, but Numerals do not attend to Numerals. For all plots, we take the mean of dataset samples to calculate the attention scores, but display only one sample on the axes for demonstration purposes.}
  \label{fig:attnpat_early_mixed_randNumerals_Months}
\end{figure*}

We discover a "similar member" detection attention head 1.5, and a "sequence member detection" attention head 4.4. Attention pattern analysis reveals that these heads detect how sequence members (as queries) attend to sequence members (as keys) of the same type, such as numerals. 
To determine if this detection only occurs if the sequence members are in sequential order, or if this occurs even if they are not, we input prompts 
with Numerals in random order but with Months in sequential order. 

In Figure \ref{fig:attnpat_early_mixed_randNumerals_Months}, we observe, for attention head 1.5, similar types attend to similar types. However, for attention head 4.4, Months attend to Months, but Numerals do not attend to Numerals, as the Numerals are not in sequential order. 
Therefore, in general, both attention heads 1.5 and 4.4 appear to detect similar token types that belong to an ordinal sequence such Numerals or Months, but attention head 4.4 acts even more specifically as an adjacent sequence member detection head. More discussion about these attention patterns are in Appendix \ref{appendix:imptHeads}. 

\textbf{Last Token Sequence Detection Head. } In Figure \ref{fig:7_11} and in Appendix \S \ref{appendix:gpt2func}, we describe results that show how attention head 7.11 acts to transfer the detected sequence information to the last token, which aids attention head 9.1 and MLP 9 in determining the successor member. 


\begin{figure}
  \centering
 \includegraphics[scale=0.35]{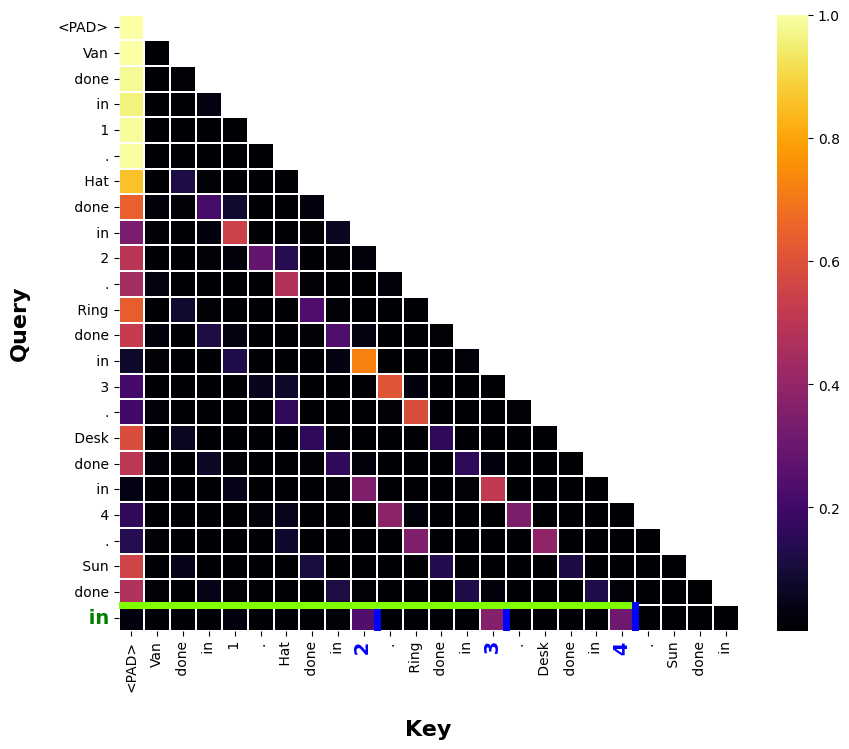}
  \caption{GPT-2 Small attention pattern for Head 7.11. At the last token, head 7.11 to more recent sequence members than earlier sequence members. We also recognize an offset pattern on the diagonals of this heatmap, indicating that it also functions as a previous token head. }
  \label{fig:7_11}
\end{figure}

\begin{figure}
  \centering
 \includegraphics[scale=0.35]{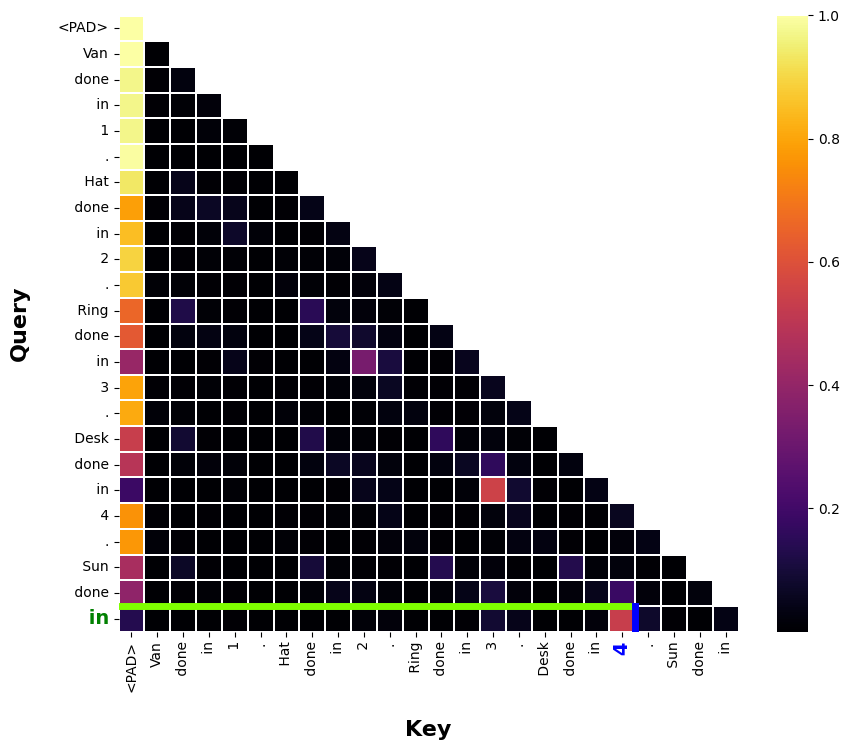}
  \caption{ This GPT-2 Small attention pattern for Attention Head 9.1 shows that for the last token, the component pays strong attention to only the most recent sequence member. }
  \label{fig:9_1}
\end{figure}

\subsection{Successor Components}

Figure \ref{fig:all_no_qkv_circ} shows that attention head 9.1 receives information from both attention heads 4.4 and 7.11. Attention head 9.1, shown in Figure \ref{fig:9_1}, pays strong attention to only the last member of the sequence, and it appears to attend even stronger to the last member than attention head 7.11, as shown in Figure \ref{fig:7_11}.
Attention head 9.1 has a successor score of 87.37 for a dataset of numerals from 1 to 97; more details on these scores is given in Appendix \ref{app:succeheads}. These results suggest that attention head 9.1 is a successor head \cite{gould2023successor}.
\footnote{Successor heads were found concurrently via independent discovery by \cite{gould2023successor}. }

\textbf{Succession via MLP}
For most samples of all sequence types, logit lens reveals that the model does not predict the correct answer before MLP 9. However, after the information is processed through MLP 9, the model outputs the next sequence member. These findings suggests that MLP 9 is largely responsible for finding the next sequence member, more so than attention head 9.1, which may just be boosting information and/or acting as a backup component. Logit lens results are in Appendix \S \ref{appendix:MLP}.

%% file: texs/7_llama2_seqcont.tex
\section{Interpreting Sequence Continutation in Llama-2-7B}
\label{sec:llamaFunc}



We apply iterative pruning using datasets of sequences (without non-sequence words in between) to Llama-2-7B, which has 1024 attention heads and 32 MLPs, to find important attention heads \textbf{5.25, 16.0, and 20.17} \footnote{For Llama-2-7B, we define a component as \emph{important} is their ablation from a task's circuit causes an average performance drop of at least 12\% for all tasks}. 
Specifically, attention head 5.25 is found to be a sequence detection head like GPT-2 Small's attention head 4.4, and attention head 20.17 is found to be a successor head like GPT-2 Small's attention head 9.1.
Given that we can find analogues of the sub-circuit's components described in \S \ref{subcirc_hypo} from one model to another, we show that this sub-circuit type appears in more than one model \footnote{Future studies can examine if this appears in even more models.}. Appendix \ref{app:llama_expms} details the setup of these experiments.

\textbf{Important Components.} 
As the full circuits' component sets are between 80 to 100 nodes and are too large to include in this paper, we only discuss important components. 
Unlike in GPT-2 Small, for Llama-2-7B we find most of the MLPs to be necessary to be able to perform sequence continuation. 
Table \ref{circHeadsDrop_llama} shows the drop in task performance for Llama-2-7B when an important attention head is removed from each respective task's circuit.

\begin{table*}
    \centering
    \caption{Drop in task performance for Llama-2-7B when an important attention head is removed from each sequence continuation task's circuit.}
    \label{circHeadsDrop_llama}
    \vspace{0.5em}
    \begin{tabular}{c|ccc}
        \toprule
        Important Head & Numerals & NumWords & Months \\
        \midrule
        5.25 & -29.24\% & -21.32\% & -12.87\% \\ 
        16.0 & -27.02\% & -14.18\% & -13.74\% \\ 
        20.17 & -56.78\% & -40.03\% & -27.26\% \\ 
        \bottomrule
    \end{tabular}
\end{table*}

\begin{figure}
  \centering
 \includegraphics[scale=0.3]{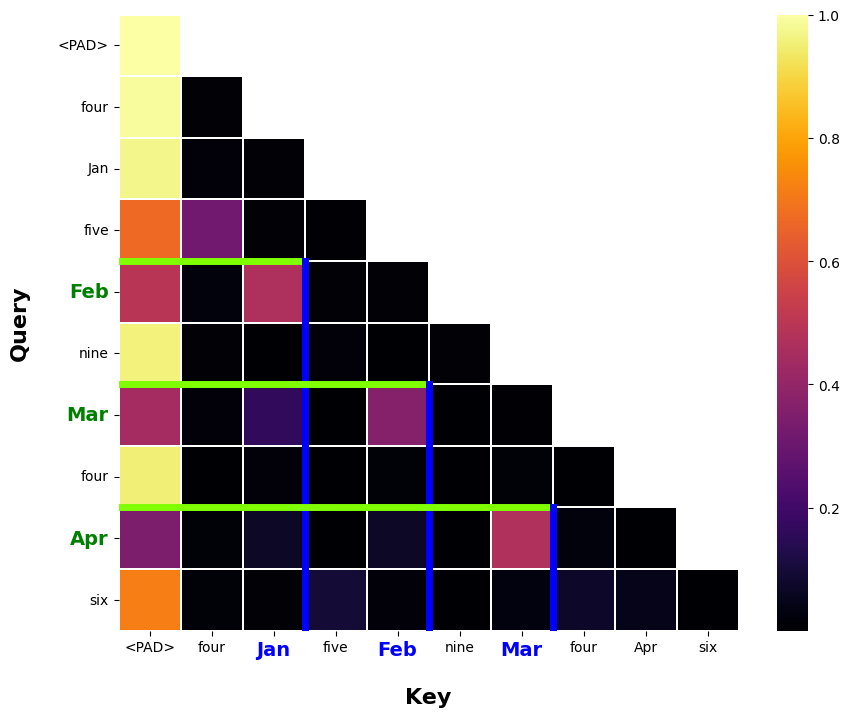}
  \caption{ The attention pattern of attention head 5.25 in Llama-2-7b resembles the attention pattern of GPT-2 Small's attention head 4.4 in Figure \ref{fig:attnpat_early_mixed_randNumerals_Months}(b), indicating they have similar functionality as sequence member detection heads. }
  \label{fig:5_25_llama}
\end{figure}

\begin{figure}
  \centering
 \includegraphics[scale=0.35]{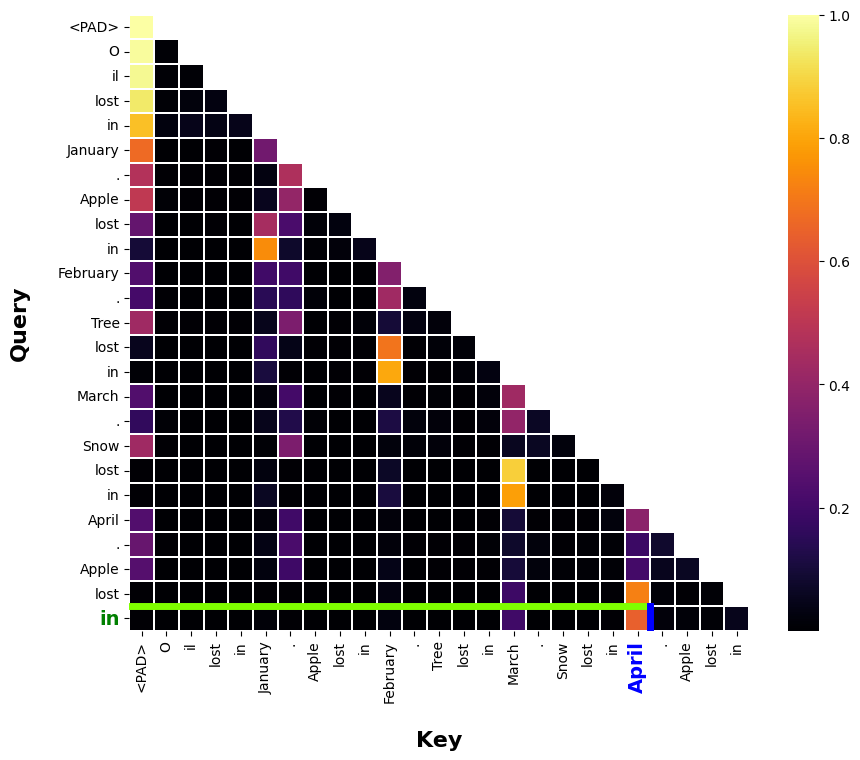}
  \caption{Llama-2-7B attention pattern for Head 20.17, bearing resemblence to the attention pattern of GPT-2 Small's Head 9.1 in Figure \ref{fig:9_1}, showing that for the last token, the component pays strong attention to only the most recent sequence member.}
  \label{fig:20_17}
\end{figure}

\textbf{Attention Patterns. } As shown in Figure \ref{fig:5_25_llama}, Llama-2-7B's attention head 5.25 appears to be a sequence detection head similar to GPT-2 Small's attention head 4.4. Moreover, it is not just a similar member detection head like GPT-2 Small's attention head 1.5; when sequence members are out-of-order, the component does not detect these members. But when the members of any of the 3 tasks are in-order, it is able to detect them. 
Likewise, in Figure \ref{fig:20_17}, Llama-2-7B's attention head 20.17 appears to be a successor head similar to GPT-2 Small's attention head 9.1. We discuss attention head 16.0 in Appendix \ref{sec:llamaFunc}.

\textbf{OV Scores. } Attention head 20.17 has a higher than average successor head score of 33.3\%, compared to the average score of 0.46\%. It is one of only twelve attention heads (out of 1024) which detects the "next" token of a sequence.

%% file: texs/8_math_behavior.tex

\section{Ablating Sequence Continuation Components on Math-Related Prompts}
\label{sec:wordprobs}

We examine how ablating sequence continuation components affects the generated output for Llama-2-7B on various math-related prompts. This demonstrates how these circuits are not just specific to a narrow range of sequence continuation tasks, but have effects on a wide range of mathematical reasoning tasks. 
We observe the ablation effects on the following prompt types: (1) Additive sequences for intervals >1 such as "2 4 6", and (2) Arithmetic. In Table \ref{ablationPromptsFull}, we provide a summary of these results. More results, including those for natural language math reasoning prompts and Spanish language sequences, are in Appendix \ref{appendix:natlangprompts}. 



We measure how well the model correctly performs completions on several prompt types after ablating component sets of our discovered sequence continuation circuits. Given a set of prompts, we define the \textbf{Percentage Destroyed after Ablation} as the percentage of prompts which do not complete correctly after ablation.
We compare our component set ablations with random component ablations which have no overlap with the intersection of all three component sets. We use 50 randomly chosen component sets for every prompt, and take their mean score. 
Appendix \ref{appendix:natlangprompts} discusses details of this ablation procedure.
\textbf{On intervaled sequences. } As shown in Table \ref{ablationPromptsFull}, for intervaled sequences of +2, +3, +10, and +100, we discover that ablating an attention head component set for each sequence continuation task destroys the model's completion ability for 96-100\% of cases. 
In contrast, using random component ablations only destroys around 3-28\% of cases. 



\textbf{On arithmetic. } 
Ablating sequence continuation head sets destroy the model's ability to perform addition and subtraction.
This suggests that sequence continuation, addition, and subtraction tasks share circuits. 
As shown in Table \ref{ablationPromptsFull}, we find that ablating the sequence continuation component heads destroy the ability to perform double-digit arithmetic for 92-100\% of our experiments, while randomly selected ablated heads only do so approximately 20-33\% of the time.

\begin{table*}
    \centering
    \caption{Percentage Destroyed after Ablation. Percentage Destroyed denotes the percentage of prompts with incorrect completion after ablation of the component sets of that column; \textbf{the higher the score, the more effective the ablation}. Each prompt type uses 50 prompts. Results for more prompt types are given in Table \ref{ablationPromptsFull_v2}.} 
    \label{ablationPromptsFull}
    \vspace{0.5em}
    \begin{tabular}{p{4cm}|p{2cm}|p{2cm}|p{2cm}|p{2cm}}
        Prompt Type & Numerals & NumWords & Months & Random 100 \\ 
        \midrule
        +1 intervals & 96\% & 100\% & 98\% & 3\% \\
        \midrule
        +2 intervals & 96\% & 100\% & 100\% & 7.7\% \\ 
        \midrule
        +3 intervals & 100\% & 100\% & 100\% & 28\% \\ 
        \midrule
        +10 intervals & 96\% & 100\% & 100\% & 4\% \\ 
        \midrule
        +100 intervals & 100\% & 100\% & 100\% & 8.2\% \\ 
        \midrule
        Double-Digit Addition & 98\% & 100\% & 100\% & 32.8\% \\ 
        \midrule
        Double-Digit Subtraction & 96\% & 92\% & 94\% & 20\% \\ 
        \midrule
    \end{tabular}
\end{table*}


%% file: texs/conclusion.tex
\section{Conclusion}
\label{sec:conclusion}


Understanding the inner workings of neural networks is essential for fostering alignment and safety. In this paper, we identify that across similar sequence continuation tasks, there exist shared sub-circuits that exhibit similar functionality. Specifically, we find sequence token detection heads and components associated with next sequence outputs. The aim of this work is to advance our understanding of how transformers leverage shared computational structures across similar tasks. By locating and comparing these circuits across models, we hope to gain insight into semantic representations of abstract concepts, which may provide evidence for universally converging hierarchical associations across models \citep{huh2024platonic, park2024geometry, templeton2024scaling}.

\section{Limitations}

As the research topic of our work is relatively new, the aim of this paper is to first investigate shared circuits for simple tasks. This way, later work may build upon it to look for shared circuits for more complex tasks that are more important for AI safety. We discuss limitations of this paper in this section.


\textbf{Methodology Assumptions. }  As mechanistic interpretability is an early research field, there may be methodological assumptions that these findings depend on which may be subject to change as the research field progresses.

\textbf{Dataset Size. } As there are only twelve months, the number of possible continuing sequences was limited. Additionally, even if months were not used, GPT-2 also has limited prediction ability for number word sequences, as detailed in Appendix \ref{appendix:dataset}. However, our dataset size for the months continuation task fully captures all of the months; in contrast, a small dataset size brings more issues when it does not capture all of the true distribution.



\textbf{Task Complexity. }  
We performed preliminary experiments on intervaled sequences. Future work can expand on this area, and also study Fibonacci-like sequences, alternating sequences, and determine if the model has diverging calculations for +2 vs x2 sequences. The aim is to generalize from simple tasks to evidence that general shared tasks models can be interpreted. Additionally, future work can test on more varied prompt types, such as more math reasoning prompts.

\textbf{Comparison Statistics. }  We have shown that there are effects from ablating these sequence continuation circuit components on the model performance for various prompt types. More rigorous statistical tests can improve these studies.

\textbf{Future Work. } Future work can build upon this paper by tackling open questions related to feature-level analysis of sequence continuation tasks, and by examining components exclusive to certain tasks to see if they handle mapping between abstract representations to specific tasks. Another research problem involves analyzing the effects of model editing on shared, entangled circuits. Approaches to this include quantifying the relationship between circuit entanglement and editing impact which may be done via embedding space projection \citep{dar2022analyzing}, modifying a sub-circuit used for a sub-task $S$ and observing if the ability to recognize $S$ in multiple tasks is destroyed, and utilizing methods such as model steering to edit a task $S$ to perform a similar task $S'$ \cite{turner2023activation, merullo2023circuit}.


\section*{Acknowledgements}
We are grateful to Torr Vision Group (TVG) and Apart Lab members for feedback on the previous version, specially, Jishnu Mukhoti, Luke Marks,  Michelle Lo and Ashkan Khakzar for comments and feedback and Clement Neo for assistance with Figure 1.

%% file: texs/apdx_expm_setup.tex
\section{Computational Resources and Packages}

The code for the experiments was written in Python, utilizing the TransformerLens package, and were run on an A100.

\section{GPT-2 Experimental Setup Details}
\label{appendix:dataset}


\textbf{Circuits Variations. } We observe that there are multiple circuits, with slight variations between them, that have similar performances for the same task. 
However, we find that important heads are often found in most circuits, regardless of the method, metric or dataset choices. Thus, we focus more on the "big picture" comparison of scores and on the most important heads, and less on the exact variations between scores or the less important heads.

\textbf{Dataset Generation Procedure. } To generate each sample, we place each sequence within a specific template that is generated from an abstract template. For instance, the abstract template of '<name 1> born in <seq mem A>. <name 2> born in <seq mem B>. <name 3> born in'
can fill in names with (Kyle, Anthony, Madison) to make a specific template. Then, the specific template can be filled in with sequence members (February, March) to create the sample 'Kyle was born in February. Anthony was born in March. Madison was born in'. We generate 1024 samples from each of three abstract templates, where each sequence (eg. 1 2 3, or 8 9 10) is represented the same number of times, for a total of 1536 samples per task. We use single tokens for all tokens in each sample. We choose samples such that the model outputs the correct answer with at least twice as high probability as the incorrect answer's probability. Each specific template must also meet these conditions for all sequences (eg. must work for 1 2 3, 8 9 10, and two three four); else, it is not used. Each template is represented in equal proportion.

We use the same templates for all tasks. For example: given the sample for the months task "Ham was bought in February. Egg was bought in March. Bread was bought in April. Steak was bought in”, we use the same non-sequence tokens to make a sample for the digits task: "Ham was bought in 2. Egg was bought in 3. Bread was bought in 4. Steak was bought in”.

The three templates we used are: <name> born in, <item> lost in, <item> done in. We choose from a set of 136 names and 100 items.

Originally, we use the token "was" in our samples (eg. "Steak was sold in March.") However, we find that the prediction outcomes are largely the same whether we included "was" or not. Thus, although "was" would make the sentences sound more natural to a human, we choose to omit it. Additionally, this allows to reduce the memory usage while running in Colab.

\textbf{Random Words vs Meaningful Sentences. } We find that using random words as non-sequence tokens could also allow the model to sometimes predict the next sequence member correctly. However, this did not always occur; thus, we choose to use semantically meaningful templates instead.

\textbf{Sequence Member Input Positions. } We did not construct samples such that there are different intervals between the sequence members, placing them at different positions in the input (eg. "1 2 house fork 3" or "1 house fork 2 3"), because we want the model to be able to predict the next sequence member with high probability. Thus, we give it an in-context pattern where after every random word, it should predict a sequence member.

\textbf{Sequence Length. } We find that using sequences with four members allows the model to consistently obtain high probability predictions for the correct answer for all three tasks. For continuing sequences without non-sequence members, four members is usually enough to obtain a correct token probability of around 90\% or more for the three tasks, within a certain range (eg. not above twenty for number words for GPT-2). 

\textbf{Model Sequence Continuation Abilities. } For number words, as GPT-2 Small does not seem to be able to continue number word sequences higher than twenty, even when giving it the starting prefix with and without hyphens (eg. twenty or twenty- for twenty-one). We add a space in front of each number word as without the space in front, the model tokenizer would break some words greater than ten into more than one token (eg. eleven into two tokens, and seventeen into three tokens), while we aim for all our samples in a dataset to have the same number of tokens. Similarly, for digit sequences there are cases where it would break the answer into multiple tokens (eg. in the 500-600 digit range, sometimes the next token predicted would be "5", and sometimes it would be "524"). 

\textbf{Corrupted Dataset Details. }  We ensure that our randomly chosen sequence does not contain any elements in sequence order for the last two elements of the input, as if the last two elements are not sequential, sequence continuation cannot successfully occur.
We also test variations of several corruptions other than randomly chosen tokens of a similar sequence type, such as repeats and permutations.
Overall, the most important components remain the same regardless of the ablation dataset and metric choices.

\textbf{Other Task Datasets. } We also look for similarities between other types of tasks, such as decreasing sequences, greater-than sequences, and alphabet sequences. However, while there are a few shared circuit overlaps between these tasks and three main tasks of this paper, there are more dissimilarities. Thus, we mainly focus on the similarities of the three tasks of this paper.

\subsection{IOI Circuit. } The IOI circuit in Table \ref{circComp} uses all MLPs and the following heads (shown in (L,H) format): 

(0, 1), (0, 10), (2, 2), (3, 0), (4, 11), (5, 5), (5, 8), (5, 9), (6, 9), (7, 3), (7, 9), (8, 6), (8, 10), (9, 0), (9, 6), (9, 7), (9, 9), 
(10, 0), (10, 1), (10, 2), (10, 6), (10, 7), (10, 10), (11, 2), (11, 9), (11, 10)




\section{Llama-2 Experimental Setup Details}
\label{app:llama_expms}

\textbf{Dataset. } For Llama-2, we ran iterative node pruning on samples consisting only of sequence members, and ran attention patterns on samples consisting of both sequence and non-sequence members. This is because for Llama-2, the main focus was on finding similar components between the sequence continuation sub-circuits of GPT-2 and Llama-2, rather than finding more minimal circuits, which would be beyond the computational resources we allocated. 
Additionally, the numbers 10 to 12 are represented as two tokens (eg. '10' is '1' and '0') in Llama-2-7b; as we evaluate only on prompts of the same length and use single token answers, we only use numbers 1 to 9 in the Numerals dataset.

\textbf{Iterative Methods.} Given that Llama-2 is larger than GPT-2, we only use on backwards sweep, which we found to be enough for finding the most important components. We also try modifying the iterative component ablation method to first ablate all the heads of an attention layer to see if this makes the model performance perform below the threshold; if it does, we remove all the heads of that attention layer from the candidate circuit. While this method variation is faster if many attention layers are unnecessary, it has the downside in which removing certain unimportant heads may increase performance, which may, in summation, offset the performance degrading abilities of important heads. This means that there may be some important heads which may be missed by this method. However, we found that the most important attention heads were not missed by this method, and the end results were often comparable to the original version of our iterative methods. The results in this paper do not use this modification.

%% file: texs/apdx_gpt_results.tex
\section{Iterative Path Patching for Edges}
\label{edge_patching}

\textbf{Path patching} is a different type of patching that allows for a more precise analysis of an intervention's effect on a particular path \cite{goldowskydill2023localizing}. It can be performed by ablating component interactions, measuring the effect of one component on another. 

After finding circuit nodes, we utilize path patching to obtain interactions (edges) between them. Edges denote nodes with high effects on other nodes \footnote{As the residual stream allows for indirect effects, edges may be between components at non-adjacent layers,}. We apply a form of iterative path patching which works backwards from the last layers by finding earlier components that affect them. First, we ablate the nodes pruned from iterative node pruning. Then, we ablate one candidate edge of the unablated nodes at a time. Using the same order as the backward sweep, we take a node as a \emph{receiver} and find the \emph{sender} nodes that have an important effect on it.
If patching the effect of sender A on receiver B causes the model performance to fall below threshold $T_e$, the edge is kept; else, it is removed. 

For example, if node pruning found a circuit that obtains a 85\% score above $T_n = 80\%$, we now measure which circuits with the ablated nodes and the ablated candidate edge still have performance above $T_e = 80\%$ \footnote{The edge pruning threshold $T_e$ may be the same or different as the node pruning threshold $T_n$.}.
Performing node pruning before edge pruning filters out many nodes, reducing the number of edges to check.
After edge pruning, nodes without edges are removed. This method has similarities to the path patching used by \cite{hanna2023does}, but with several differences, such as using our performance metric as a threshold.

\section{Functionality Method Details }
\label{appendix:fmethod}

\textbf{Component Output Scores Details. } To continue from Section \S \ref{sec:methods}, we employ the heads' output projection ($\mathbf{OV}$) matrices to examine the attention head outputs written to the residual stream by the OV circuit. For example, we can check if a head is copying tokens, a behavior introduced as \emph{copy scores} by \cite{wang2022interpretability}. Copy scores measure how well a head reproduces a token from the input.
Similar to \cite{gould2023successor}, who analyzed successor heads in greater detail than our paper, we modify this method to obtain the \emph{component output score}, or \textbf{OV score}, which follows a similar principle but measures how many prompts have a \emph{keyword} token are in the output. To calculate these scores, we multiply the state of the residual stream after the first MLP layer at the last token with the OV matrix of the attention head of interest. This result is unembedded and layer normalized to get logits. If the keyword is in the top-5 of these logits, +1 is added to the score. Finally, we divide the total score by the total number of keywords across all prompts to obtain a percentage which we call the OV score. Specifically, we use the keyword as the integer $I+1$, given integer $I$ as the last token in a sequence, to obtain an OV score known as the \textbf{successor score}.
In this paper, we use all sequence members of the prompt as keywords.

\section{GPT-2 Small Connectivity Details}
\label{appendix:gpt2conn}

As seen in both Figure \ref{fig:full_all_no_qkv_circ} and in Table \ref{allHeadDropsCircuit} in Appendix \ref{appendix:indiv_circ}, in which only head 5.0 of the Numerals circuit is not part of the Number Words circuit, the Numerals circuit is nearly a subset of the Number Words circuit. This suggests that the Number Words circuit uses the Numerals circuit as a sub-circuit, but requires additional components to make accurate predictions.

\renewcommand{\arraystretch}{1.5}
\begin{table*}
\centering
\caption{Performance Scores for Figure \ref{fig:full_all_no_qkv_circ} Circuits' Components (cols) run on Similar Tasks (rows) }
\label{circComp}  
\vspace{0.5em} 
\begin{tabular}{c|cc cc|c}
\toprule
  & Numerals Task & NumWords Task & Months Task \\
\midrule
Numerals Circuit & 81.01\% &  48.41\% & 113.52\% \\
Number Words Circuit & 87.35\% & 81.11\% & 103.64\% \\
Months Circuit & 43.74\% &  32.36\% & 80.30\% \\
IOI Circuit & -6.70\% &  -15.82\% & -9.20\% \\
\bottomrule
\end{tabular}
\end{table*}

%

In Table \ref{circComp}, we compare every task's circuit with other similar tasks, isolating each circuit by resampling ablation on non-circuit components. First, we observe that in general, the model cannot perform well on these tasks for non-sequence-task circuits. For instance, we show that the model has negative performance for all tasks when run on the IOI circuit. The negative values mean that the $(L_C) - (L_I) < 0$ in the ablated circuit, indicating bad performance.

We observe that for the Numerals task, the model performs better on the Number Words circuit than the Numerals circuit, which may be because the Numerals circuit is nearly a sub-circuit of the Number Words circuit. It is possible to find a Numerals circuit with higher performance by setting the threshold higher. However, this paper's pruning methods attempt to find minimal circuits with only necessary components above a certain threshold; they do not seek to find the circuit with the most optimal performance \footnote{One can obtain circuits with >100\% performance by setting the threshold to be 100.}.

For the Number Words task, the model only performs well with the Number Words circuit, as this task may require more components than the other two. On the other hand, for the Months task, the model performs even better than the unblated circuit for all sequence-task circuits, indicating that this task may not require as many components as the other two. Due to components such as inhibition heads \citep{wang2022interpretability}, ablating certain heads may allow the model to perform better for specific tasks, though may hurt its ability on other tasks. Overall, these results show that these tasks do not use the exact same circuit,  but may have partially good performance on other sequence task's circuits due to shared sub-circuit(s).

\section{GPT-2 Small Functionality Details}
\label{appendix:gpt2func}


\textbf{Duplicate Head} Head 0.1 was noted to be a Duplicate Token Head by \cite{wang2022interpretability}, in which it recognizes repeating patterns. As we did not note that 0.1 had any effects on sequence members in particular, given our non-sequence token patterns, it is likely that 0.1 is recognizing all repeating patterns in general, which is prevalent in our dataset. Though it plays an important role for this sub-task, it does not appear specific to sequence continuation.

\subsection{Last Token Sequence Detection Head} 


In Figure \ref{fig:all_no_qkv_circ}, there is an edge from heads 1.5 and 4.4 to 7.11, showing 7.11 obtaining sequence token information from earlier heads. Then, we observe in Figure \ref{fig:7_11} that for head 7.11, query tokens attend to its previous key tokens, indicating 7.11 acts like a "Previous Token" head. Noticeably, at the last query token, the strongest attention appears to be from the non-sequence tokens to the sequence tokens. This head may "ordering" identified sequence tokens to send to the last token, or it may be detecting the pattern at which token the model should predict the next member of the identified sequence (eg. after each non-number token, the next member of the number sequence often follows.)

\subsection{Successor Head Scores} 
\label{app:succeheads}

To check that head 9.1 outputs next sequence tokens, we study its component output scores. Table \ref{table:top_5_output_h9_1} shows that given a numeral token I as input (eg. 1), head 9.1 often outputs a token I+1 or higher (eg. 2).
For numerals between 1 and 97, its successor score is 87.37\%, while its copy score is 59\%. 
We also note that the successor scores of most heads are low, with an average of 3.29\%, 
and that head 9.1 has the highest successor score.
Thus, it seems to function as a "successor head". 

This is reinforced by the successor score for number words, which is 90.63\%, while the average for all attention heads is 2.97\%.
Although head 9.1 does not appear to output months given any month token, we observe something peculiar: 9.1 is the \emph{only} head that will output the next rank given a month (eg. given "February", output "third", and its "next rank given month" score is 31.25\%. This appears to be related to how months can be mapped onto ranks.


\begin{table}[h!]
  \centering
  \renewcommand{\arraystretch}{1.5} 
  \caption{The top-3 tokens output tokens after OV Unembedding head 9.1 for several input tokens. }
  \vspace{0.5em} 
  \label{table:top_5_output_h9_1}
  \begin{tabular}{c|c}
    \toprule
    \textbf{Token} & \textbf{Top-3 Tokens after Unembed} \\
    \midrule
    `78' & ' 79', `80', `81' \\ 
    `six' & ' seventh', ' eighth', ' seven' \\ 
    `August' & `ighth', `eighth', `ninth' \\ 
    \bottomrule
  \end{tabular}
\end{table}

%% file: texs/apdx_llama_results.tex
\section{Llama-2-7B Functionality Details}
\label{appendix:llama2func}



\begin{figure}
  \centering
 \includegraphics[scale=0.35]{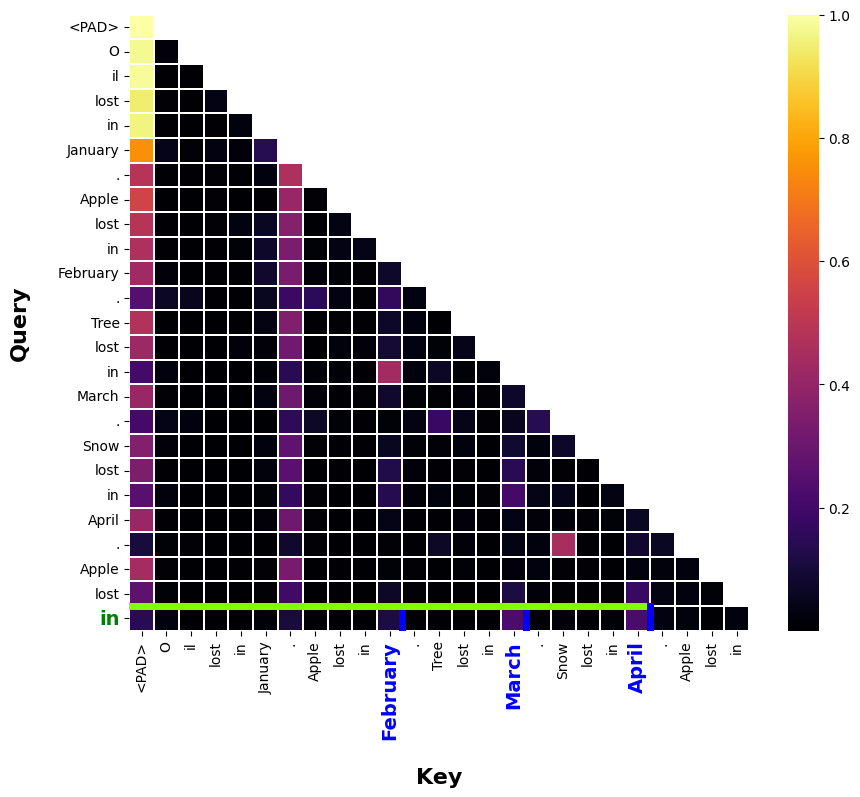}
  \caption{Llama-2-7B attention pattern for Head 16.0. Tokens appears to attend to the last period of the first instance of the pattern (<object> lost in Month.) However, it is not clear from this plot what the role of this head is. }
  \label{fig:16_0}
\end{figure}

\textbf{Other Attention Patterns.} Figure \ref{fig:16_0} shows attention patterns attention 16.0
It may a slight resemblence to the "last token sequence detection head" 7.11 in GPT-2 Small, but it is not clear from this plot what the role of this head is. 



\textbf{Interval-K Sequences: Attention Patterns} 
We apply the iterative ablation methods to intervaled sequences such as additive +2 (eg. "2 4 6"). We discover that heads 16.0 and 20.17 are within these components sets, and that head 6.11 is considered especially important. Future studies can delve more into this area.

The attention patterns for the important attention heads 5.25 and 20.17 on the Interval-K tasks, where $K \geq 1$, are similar to those for Interval-1. For intervals that are not a multiple of 10 and for numbers with more than one digit, the last digit of a member attends to the last digit of a previous member. For intervals that are a multiple of 10, these digits attend to the differing number (eg. the '3' in 300 attends to the '2' in 200). 









%% file: texs/apdx_abl_gen.tex
\section{Ablation on Math-Related Prompts} 
\label{appendix:natlangprompts}

\textbf{Experimental Setup. } 
We run ablation experiments on various math-related prompt types only with Llama-2-7B as GPT-2 Small is limited in its ability to correctly complete many math-related prompt types.
Zero ablation is performed for these experiments, in which the components being knocked out are replaced with an activation of 0, rather than mean activations from a corrupted dataset. This is because for certain prompts like for natural language, it is not always clear what type of corruption to choose, so zero ablation is easier to work with.

For sequence continuation prompt types, we use sequences of length-3, showing that though we used length-4 data samples to find important components, that our findings generalize to sequences of different lengths. For arithmetic prompt types, we use 50 different arithmetic expressions where the correct answer is greater or equal to 0.

For random ablation on intervaled sequences and arithmetic, a sequence continuation component set for a task is around 80-90 attention heads out 1024 total attention heads. Thus, we choose to ablate 100 attention heads. For the randomly selected component sets for all prompt types, we do not select attention heads that belong to the the intersection set of attention heads from the 3 circuit found for the three sequence continuation tasks. In other words, the Numerals circuit component set has 86 attention heads, and taking the intersection of this component set with the component sets from the Number Words and Months circuits yields an intersection set of 16 attention heads. None of these 16 heads are not selected when obtaining attention heads for the randomly selected component sets.

Overall, we note that these tasks are not solely dependent on the sequence continuation component sets we find, which means random ablations may choose components that are crucial for the task. For instance, the random ablation score for the arithmetic may likely be lower, as some of its runs may be ablating heads that are involved in arithmetic. 





\textbf{Math-Related Word Problems. }
The following experiments provide some evidence that the sequence continuation circuits are not specific to just sequence continuation, but can extend to prompts involving sequence continuation reasoning. More experiments on larger datasets and more prompts can determine if statement holds in more cases. In summary, we evaluate on the following prompt types, with type (2) and (3) being prompt templates:

\begin{enumerate}
    \item "What are the months in a year? Give all of them as a list. Be concise."
    \item "If this month is X, and Y months pass, what month is it? Answer: "
    \item "If today is the Xth of (month M), what date will it be in Y days? "
\end{enumerate}


(1) \textit{"What are the months in a year? "} We observe the effects of ablating sequence continuation task circuits on the prompt, "What are the months in a year? Give all of them as a list. Be concise." To evaluate the model's answer, we allow the model to generate an output of 50 tokens.
We apply the iterative pruning algorithm to obtain component sets for the three tasks. When ablating any of the 3 sequence continuation component sets, which contain around 80 heads each (eg. the Numerals circuit has 86 heads), the model loses the ability to correctly complete the answer. In particular, ablating the Numerals attention head set will make the model stop the list at June, ablating the Number words attention head set will make the model stop the list at February, and ablating the Months attention head set will make the model state the output as "1. January 2. February 2. February 3. March 3. April 3. May 3. August 3. August", listing each item on a new line. We note that the Numerals and Number Words head sets do not particularly show strong evidence for sequence destruction, but the Months head set clearly shows the model loses the ability to correctly complete the sequence.
In contrast, choosing 86 random heads that do not overlap with the Numerals circuit component set (which has 86 heads) will allow the model to retain the correct ordering for 47 out of 50 random samples of 86 random heads. We evaluate these outputs using GPT-4.

(2) \textit{"If this month is X, and Y months pass, what month is it? Answer: "} We study ablation on prompts of the format, "If this month is X, and Y months pass, what month is it? Answer: ".
We evaluate on 43 prompts generated using this template that the model obtains the correct answer on when unablated.

We find that ablating the 3 functionally important heads yields a \textit{percentage destroyed} score of 37.21\%, while using 20 randomly sampled component sets of three attention heads yields a mean destruction score of 12.62\%. For each run, we calculate the percentage destroyed score, and take the mean over every run. 
We also find that the three functionally important attention heads 5.25, 16.0 and 20.17 may work together, because if one of them is not present, the model still retains its correct completion ability. 

To evaluate if the model obtains the correct answer, we notice that for each prompt, the model may answer in different formats. For instance, for the prompt \textcolor{blue}{"If this month is December, and 12 months pass, what month is it? Answer: "}, the model answers: \textcolor{teal}{"12 months have passed, so it is now December again"}.
However, for the prompt, \textcolor{blue}{"If this month is April, and 9 months pass, what month is it? Answer: "}, the model answers: \textcolor{teal}{"If it is April and 9 months pass, then it will be January"}. Thus, we determine if the model obtains the correct answer by extracting all the months it answers with for each prompt. Then, we manually inspect both the full answer and the extracted month to determine if the extracted month is actually consistent with the model logically stating the extracted month as the answer, or if the model is simply "mentioning" that month. We find that for the prompts we use, the model is actually logically stating the extracted month as the answer. We allow the model to generate 15 tokens in its output. We found that for most prompts, this is enough for the model to correctly obtain the right answer, as shown for a sample in Table \ref{wordProblem}.

We also found that giving an instruction such as "Be concise" or a pattern we expect the model to complete in, such as "If this month is March, and 2 months pass, what month is it? Answer: May. ", often makes the model give wrong answers. Instead, we found that just letting the model generate from the question itself would yield often the correct answer.


(3) \textit{ "If today is the Xth of (month M), what date will it be in Y days? "} 
The correct answer for the majority of prompts is given in "Month Day" format (eg. January 10th). Thus, we the model to generate 5 tokens in its output.

We use 41 prompts that the model completes correctly, and discover that we only need to ablate the 3 functionally important heads for sequence continuation- 5.25, 16.0, and 20.17- to disrupt these word prompt correct completion. In contrast, ablating just 3 heads at random often does not destroy the model's correct completion abilities. In detail, we find that ablating our three important heads destroys the correct completions for 24.4\% of prompts, while randomly ablating on 3 heads, on average over 10 randomly sampled component sets, destroys 9.2\% of prompts. We also notice that ablating the 3 functionally important heads often shifts the day by just a few days, rather than changing the month. 

\underline{Overall Analysis:} 
We note that the last two prompt types are still sensitive to ablating any 3 heads of the model, suggesting that the model may require most of the heads of the model to work properly for these tasks, or that its predictions for the right answer are not as strong (having more weight on the right answer over other tokens). The latter is likely true given that the model does not always obtain the correct answer for these prompt templates; given 100 prompts, we selected 43 and 41 prompts, respectively, for the previous two templates as these were the ones the model correctly completed on. The destruction score for the 3 heads is also closer to the random ablation score. Thus, these prompt types are relatively weaker evidence for the effect of the discovered sequence continuation circuits on math-reasoning word problems; nonetheless, there is some noticeable effect, so we have reported its results in this Appendix. Using a better (fine-tuned or larger) model that has improved reasoning skills can mitigate this issue. More randomly selected samples may also be used.

In contrast to the math-related reasoning prompts discussed above, we find that a "simpler" sequence continuation task such as "1 2 3" requires more heads- around 40 to 80- for ability destruction to occur for most prompts. This may suggest that more complex prompts are more sensitive to requiring most components be healthy, while certain simpler tasks do not require this and are only destroyed when many components are destroying, possibly suggesting that these large circuit component sets we discovered may contain many backup sub-circuits within them.

As speculation, it is possible that the model may be translating to a "common space" to calculate these answers. For instance, the model may be translating "months" to an ordinal index (eg. April to 4), adding these months as numerals (modular 12), then translating the answer back to months. This speculation is be related to the findings of \cite{gould2023successor}, which found that linear maps projecting out ordinal (eg. April to 4) and domain (eg. April to month) feature abstractions could be obtained from activation space and added together, such as adding a month feature with a 2 feature to obtain a representation that is decoded as "February" in the output space. It is also related to the work of \cite{wendler2024llamas}, which found evidence that LLMs, given foreign language input, first pivot it to English to "think/reason", and then translate the answer back to the original input language.

Lastly, these prompts show that Llama-2 can reason with circular sequences, such as knowing that "Nine months after August" is May. Circular sequences have been studied in previous mechanistic interpretability work \citep{nanda2023progress, gould2023successor}, and recent work has discovered evidence that month and day features may be organized in non-linear representations \citep{engels2024language}. Future work can expand upon these studies by understanding how non-linear features causally effect other parts of a neural network, and possibly steer groups of non-linear features to observe their downstream effects on math-related circuits.

\begin{table*}
    \centering
    \caption{Percentage Destroyed after Ablation. Percentage Destroyed denotes the percentage of prompts with incorrect completion after ablation of the component sets of that column; \textbf{the higher the score, the more effective the ablation}. Each prompt type uses 50 prompts.} 
    \label{ablationPromptsFull_v2}
    \vspace{0.5em}
    \begin{tabular}{p{4cm}|p{2cm}|p{2cm}|p{2cm}|p{2cm}}
        Prompt Type & Numerals & NumWords & Months & Random 100 \\ 
        \midrule
        Single-Digit Addition & 62\% & 46\% & 92\% & 15.6\% \\ 
        \midrule
        Single-Digit Subtraction & 72\% & 46\% & 74\% & 12.4\% \\         
    \end{tabular}
\end{table*}

\begin{table*}
    \centering
    \caption{Examples of word problem prompt outputs, comparing unablated vs ablated components generation results for Llama-2-7B. For clarity, we cut off the rest of the generation once the correct answer is given. Correct answers are in \textcolor{green}{green}, and incorrect answers are in \textcolor{red}{red}. }
    \label{wordProblem}
    \vspace{0.5em}
    \begin{tabular}{p{5cm}|p{3cm}|p{3cm}|p{3cm}}
        Prompt & Original Output & 5.25, 16.0, 20.17 Ablation Output & Random Ablation Output \\
        \midrule
        \midrule
         If this month is August, and 9 months pass, what month is it? Answer: & If it is August and 9 months pass, then it is \textcolor{green}{May}. &  If it is August and 9 months pass, then it is \textcolor{red}{November}. & 9 months after August is \textcolor{green}{May}. \\  
        \midrule   
         If today is August 19th, then in 26 days it will be & \textcolor{green}{September 14th}. &  \textcolor{red}{September 12th}. &  \textcolor{green}{September 14th}. \\           
    \end{tabular}
\end{table*}

\textbf{Spanish Words Sequence Continuation. } We looked into Llama-2-7B's capabilities for sequence continuation across various languages, but found it to be limited. For instance, the model is able to complete "uno dos tres" correctly, but will output numeral "6" for prompt "dos tres cuatro cinco" instead of "seis". Thus, we only study two prompts for Spanish number word continuation that Llama-2-7B successfully completes on: "uno dos tres" and "seis siete ocho", and two prompts for Spanish months continuation: "enero, febrero, marzo". Lastly, we find the model can also correctly complete Spanish months continuation prompts when asked in natural language question formats, and find that ablating the sequence continuation circuits also affects these inputs.

For these prompts, we discover that English and Spanish number words and months share similar important components. Thus, we find similar results for Spanish as in English and using numerals. This ties into previous interpretability work on translation between English and other languages \citep{kojima2024multilingual, wendler2024llamas}.

\textit{Spanish number word continuation. } For "uno dos tres", removing just head 20.17 or 16.0 is enough to destroy this continuation ability of the model. In comparison, on average, removing any one random head that is not either 20.17 or 16.0 will not destroy this ability.
However, "seis siete ocho" requires ablating the sequence continuation head subsets to destroy the counting ability. 

\textit{Spanish months continuation. } For the Spanish months "enero, febrero, marzo", removing the intersection of the Llama-2 sequence continuation subset is enough to disrupt the model's ability to work on continuing Spanish month sequences. 

%% file: texs/appendix.tex
\section{All Three Circuits Diagram for GPT-2 Small}
\label{appendix:full_circs}

\begin{figure*}[ht]
  \centering
  \includegraphics[scale=0.195]{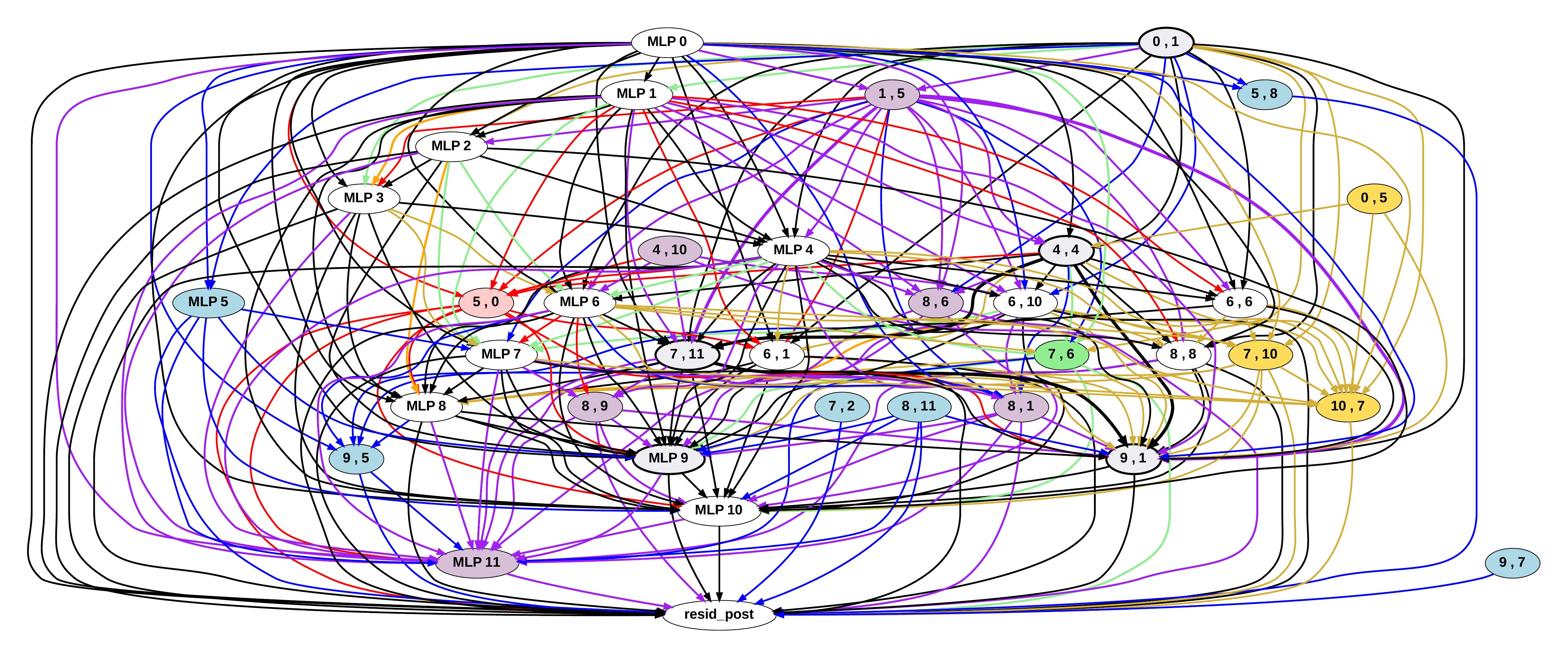}
  \caption{ A \textcolor{red}{Numerals Sequence Circuit (red)}, a \textcolor{blue}{Number Words Sequence Circuit (blue)}, a \textcolor{gold}{Months Sequence Circuit (gold)}. The overlapping sub-circuit parts are coded as follows: \textcolor{violet}{Numerals and Number Words only are in purple}, \textcolor{orange}{Numerals and Months only are in orange}, \textcolor{darkgreen}{Number Words and Months only are in green}, and \textcolor{lightgray}{All Three Tasks are in white} with \textbf{black edges}. The \textcolor{gray}{most important sub-circuit components are in gray with a \textbf{bold outline}}. Resid\_post denotes the residual stream state right before the linear unembedding to logits. }
  \label{fig:full_all_no_qkv_circ}
\end{figure*}

In Figure \ref{fig:full_all_no_qkv_circ}, we show the three circuits described in \S \ref{sec:connExpms} found via iterative node and edge ablation. Due to its large size with many edges, we place it in the Appendix rather than in the main paper.

\section{Individual Circuit Results for GPT-2 Small}
\label{appendix:indiv_circ}

\begin{figure*}[t]
  \centering
 \includegraphics[scale=0.165]{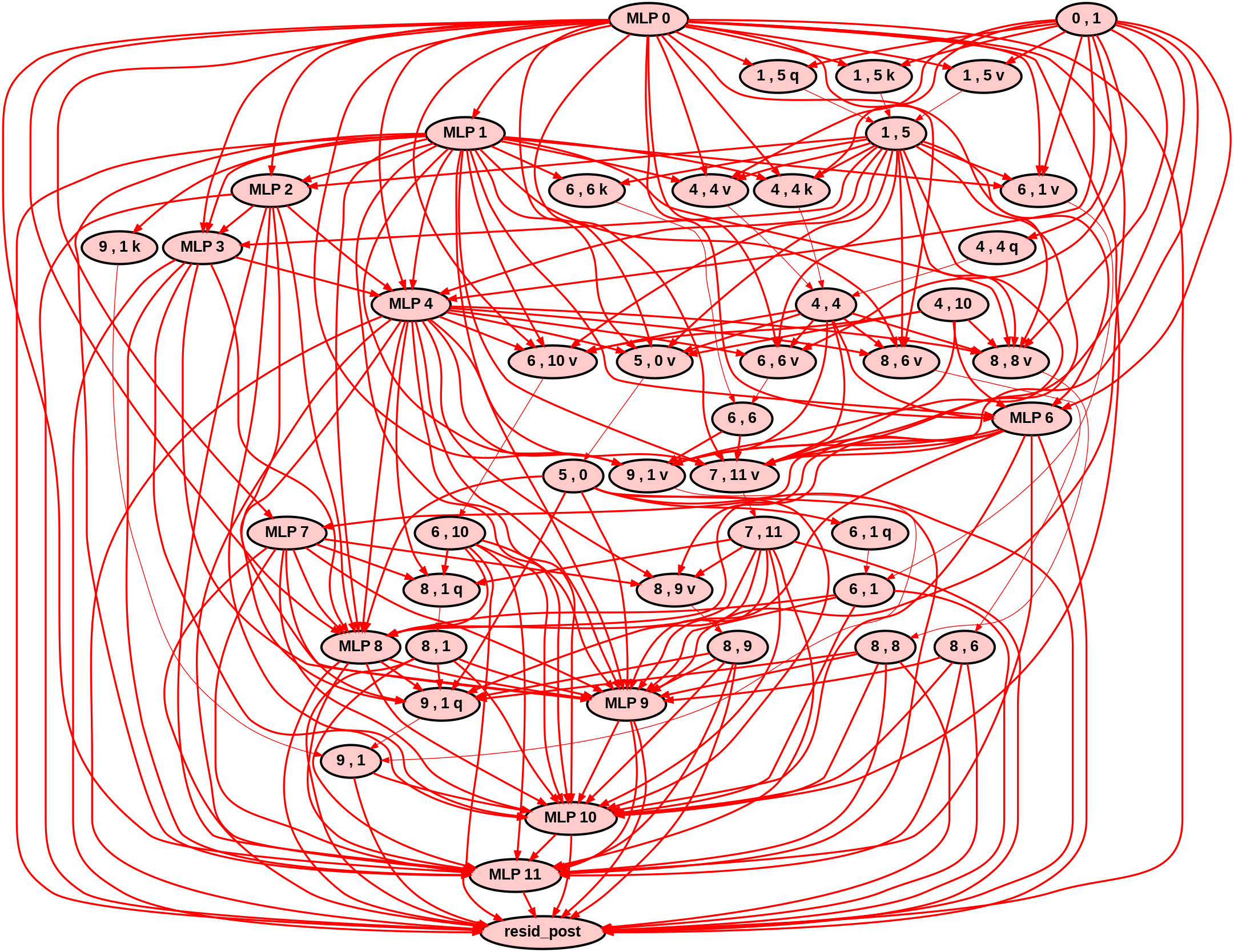}
  \caption{Numerals Circuit with Attention Head (QKV) Decomposition. }
  \label{fig:numerals_qkv_circ}
\end{figure*}

\begin{figure*}[t]
  \centering
 \includegraphics[scale=0.16]{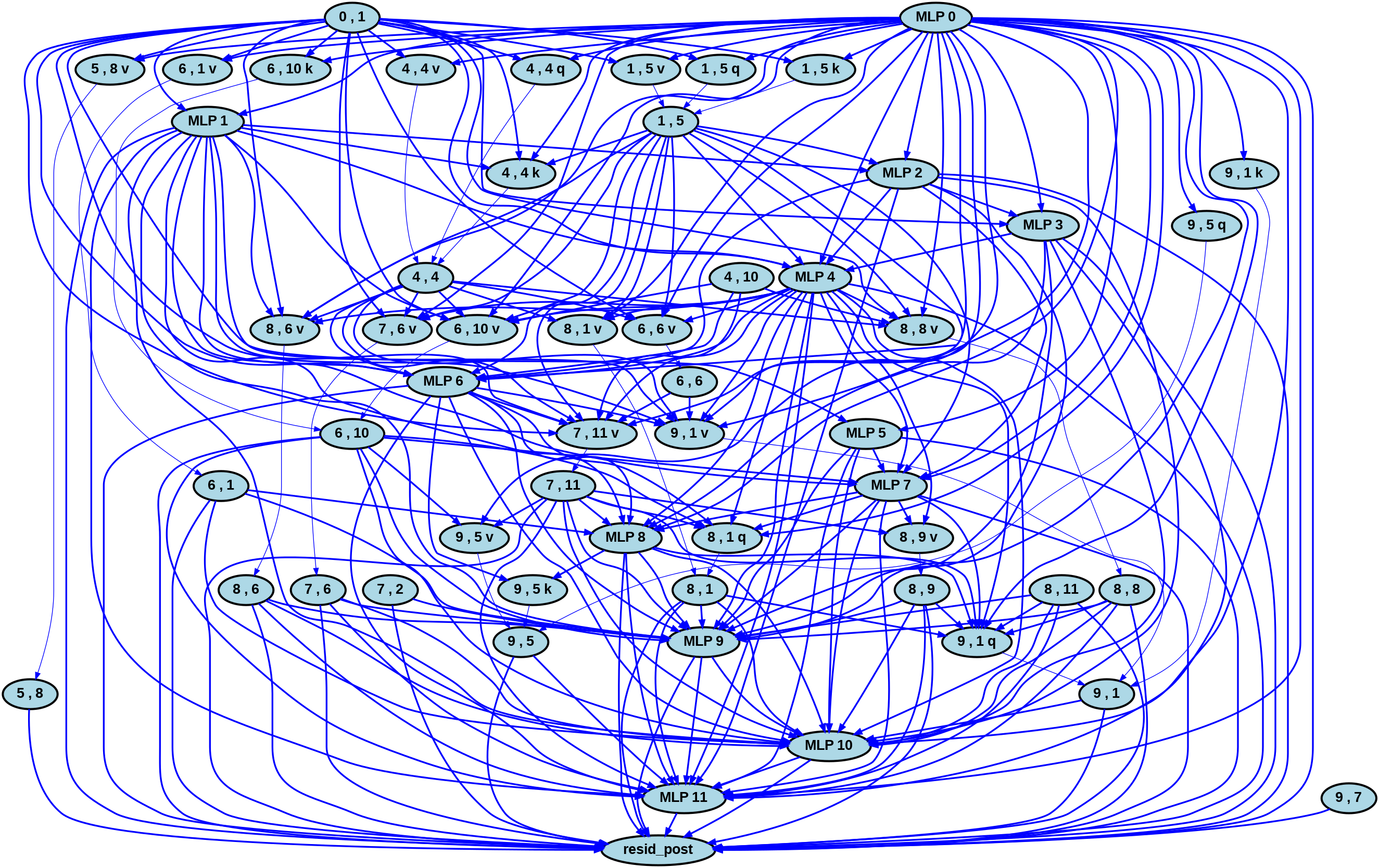}
  \caption{Number Words Circuit with Attention Head (QKV) Decomposition. }
  \label{fig:numwords_no_qkv_circ_v2}
\end{figure*}

\begin{figure*}[t]
  \centering
 \includegraphics[scale=0.163]{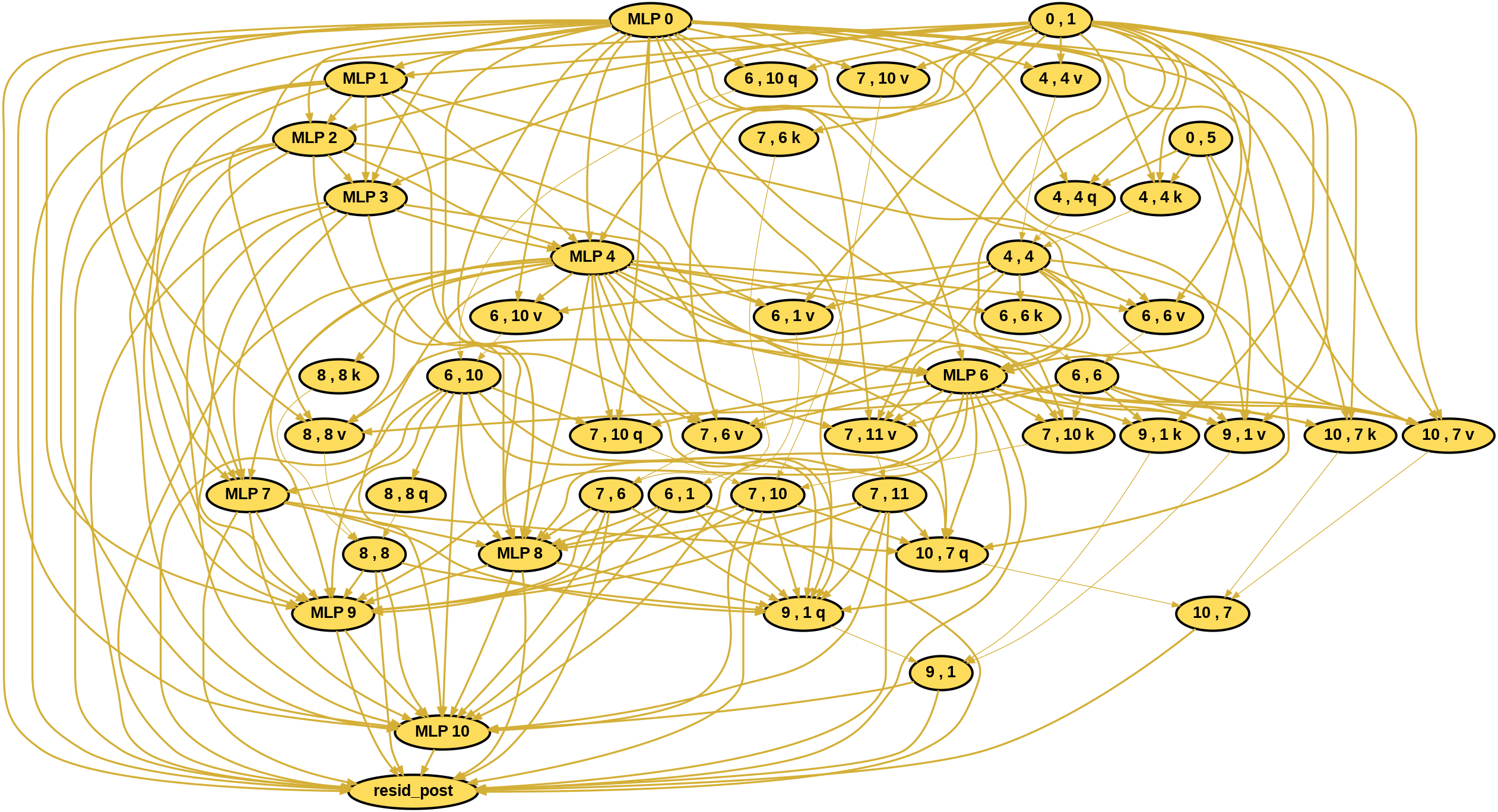}
  \caption{Months Circuit with Attention Head (QKV) Decomposition. }
  \label{fig:months_qkv_circ}
\end{figure*}

Figure \ref{fig:numerals_qkv_circ} shows a Numerals circuit, Figure \ref{fig:numwords_no_qkv_circ_v2} shows a Number Words circuit, and Figure \ref{fig:months_qkv_circ} shows a Months circuit, each with Attention Head Decomposition.
In Table \ref{allHeadDropsCircuit}, we show the result of dropping each head from the circuit shown in each of the Figures. 

In Table \ref{circHeadsDropFull}, we show the result of dropping each head from the \emph{fully unablated} circuit shown in each of the Figures. While Head 0.1 is of little importance when using the full circuit for the Numerals task with a -4.60\% performance drop when ablated, it is of very significant importance for the Number Words task, with a -91.90\% performance drop when ablated. Similar results are found for heads 4.4 and 9.1. This may occur because the model has learned multiple "backup circuits or paths" for the Numerals task, which activate when main components are ablated; it may also suggest that these heads are not important when the full circuit is present and are only important when certain components are ablated, acting as backup. The results also demonstrates that,  for the Months task, the model places different importance on the heads than for the other two tasks.
Overall, this shows that for each task, though the model re-uses many of the same important circuit parts, the importance of each part for each task varies greatly. 


\begin{table*}
    \centering
    \caption{All Head Drops from Circuits of Figure \ref{fig:all_no_qkv_circ}.}
    \label{allHeadDropsCircuit}
    \vspace{0.5em}
    \begin{tabular}{c|ccc|c}
        \toprule
        Important Head & Numerals & NumWords & Months & Average \\
        \midrule
        0.1 & -44.29\% & -78.74\% & -52.10\% & -58.38\% \\
        4.4 & -33.19\% & -34.11\% & -73.16\% & -46.82\% \\
        7.11 & -41.64\% & -44.78\% & -45.37\% & -43.93\% \\
        9.1 & -34.94\% & -27.74\% & -43.03\% & -35.24\% \\
        1.5 & -27.83\% & -18.65\% & - & -23.24\% \\
        \hline
        6.10 & -14.00\% & -24.28\% & -16.90\% & -18.39\% \\
        10.7 & - & - & -13.1\% & -13.10\% \\
        8.8 & -15.23\% & -13.21\% & -10.15\% & -12.86\% \\
        8.1 & -12.93\% & -12.61\% & - & -12.77\% \\
        8.11 & - & -10.86\% & - & -10.86\% \\
        6.6 & -7.56\% & -9.70\% & -8.93\% & -8.73\% \\
        8.6 & -11.02\% & -6.22\% & - & -8.62\% \\
        7.10 & - & - & -6.25\% & -6.25\% \\
        6.1 & -10.28\% & -4.49\% & -3.77\% & -6.18\% \\
        4.10 & -4.87\% & -5.73\% & - & -5.30\% \\
        5.8 & - & -5.15\% & - & -5.15\% \\
        5.0 & -5.02\% & - & - & -5.02\% \\
        7.6 & - & -4.96\% & -5.23\% & -5.10\% \\
        9.5 & - & -5.84\% & -3.77\% & -4.81\% \\
        0.5 & - & - & -3.79\% & -3.79\% \\
        8.9 & -4.09\% & -3.36\% & - & -3.72\% \\
        9.7 & - & -3.08\% & - & -3.08\% \\
        7.2 & - & -2.84\% & - & -2.84\% \\        
        \bottomrule
    \end{tabular}
\end{table*}

\begin{table*}
    \centering
    \caption{Drop in Task Performance when a Head is Removed from the Full, Unablated (Original) Circuit.}
    \label{circHeadsDropFull}
    \vspace{0.5em}
    \begin{tabular}{c|ccc}
        \toprule
        Important Head & Numerals & NumWords & Months \\
        \midrule
        0.1 & -4.60\% & -91.90\% & -29.89\% \\
        4.4 & -13.10\% & -52.08\% & -54.40\% \\
        7.11 & -47.21\% & -61.51\% & -46.63\% \\
        9.1 & -8.78\% & -29.93\% & -44.01\% \\
        1.5 & -14.30\% & -38.03\% & -13.15 \\
        \bottomrule
    \end{tabular}
\end{table*}

\section{MLP Analysis Details for GPT-2 Small}
\label{appendix:MLP}

In Table \ref{mlpsDropFull}, we show the performance drop for the three tasks when ablating each MLP from the full, unablated circuit. We note that MLP 0 and MLP 9 are highly important. MLP 0 may be important due to acting as a "further embedding" after the embedding layer, which embeds the tokens into latent space. 
For all numeral sequence-member only samples ("1 2 3 4" to "8 9 10 11"), we find with logit lens that the "last sequence member" (eg. for 1 2 3, this is "3") is always output at some layer between MLP 6 to MLP 8. However, after MLP 9 to the last MLP, the output is always the next sequence member. The logit lens results for the top-3 tokens at each layer for a sample with non-sequence-members is shown in Table \ref{logitLensSample}, and the logit lens results for a sample with only sequence-members is shown in Table \ref{logitLensSample_pureNumerals}. This pattern occurs in 1531 out of 1536, or 99.67\%, of the samples with non-sequence-members. The anomalies have MLP 8 predicting the correct answer of '5' or '7'.

However, for number words with only sequence-members, MLP 9's role is not so clear. In some cases, MLP 8 will output the last sequence member and MLP will output the next one. In other cases, MLP 8 will output the last sequence member as a numeral, and MLP 9 will output the next sequence member as a number word. Yet in other cases, MLP 8 will output a number word related token, such as "thousand" or "teen", and MLP 9 will output the correct answer. For one sample, "six seven eight nine", the token '10' is outputed by MLP 9, and only until MLP 11 does the output become 'ten'. Table \ref{logitLens_nw} displays a number words prompt's results.

The pattern of MLP 8 outputting a sequence member before MLP 9 outputs the next sequence member occurs in 1396 out of 1536, or 90.89\%, of samples with non-sequence-members. The main culprits where this does not occur are for sequences that have correct answers of "seven" (in which MLP 8 outputs "seven") or "ten" (in which MLP 9 outputs '10' and MLP 10 outputs 'ten'). These results suggest that the role of MLP 9 is more nuanced than simply acting as a key:value store for next sequence members. Instead, this task may be distributed across various components, with MLP 9 being one of the most important parts for this task.

For months, all the samples with only sequence-members have the last sequence member at MLP 8, and the next sequence member at MLP 9. For samples with non-sequence-members, this occurs in 1495 out of 1536, or 97.33\%, cases. The anomalies are samples that have the correct answer of "September", in which MLP 8 will output September. Table \ref{logitLens_months} shows that for the sample with only sequence-members that has the correct answer of "September", this does not occur, but strangely, MLP 0 will output "Aug" while MLPs 1 to MLP 5 will output years.
It is possible that the sequence of months is more predictable than the other sequences. This is because for numerals and number words, a sequence of numerals doesn't always result in the next one, as there can be cases in natural language where "1 2 3 4" results in "55" because it is recording counts in general, or there may be some non-linear growth. Unlike numbers, months are more constrained in a smaller range.


\begin{table}
    \centering
    \caption{Drop in Task Performance when a MLP is Removed from the Full, Unablated (Original) Circuit.}
    \label{mlpsDropFull}
    \vspace{0.5em}
    \begin{tabular}{c|ccc}
        \toprule
        MLP & Numerals & NumWords & Months \\
        \midrule
        0 & -62.58\% & -95.98\% & -84.80\% \\
        1 & -9.28\% & -34.71\% & -8.30\% \\
        2 & -2.68\% & -20.18\% & -16.40\% \\
        3 & -2.67\% & -18.19\% & -9.33\% \\
        4 & -14.19\% & -49.24\% & -23.88\% \\
        5 & -12.64\% & -25.16\% & 6.42\% \\
        6 & -15.83\% & -33.46\% & -10.22\% \\
        7 & -11.90\% & -29.71\% & -19.42\% \\
        8 & -25.19\% & -43.17\% & -41.33\% \\
        9 & -71.33\% & -84.10\% & -83.97\% \\
        10 & -32.71\% & -42.09\% & -32.53\% \\
        11 & -21.16\% & -24.97\% & -19.50\% \\
        \bottomrule
    \end{tabular}
\end{table}


\begin{table}
    \centering
    \caption{Logit Lens- "Anne born in 2. Chelsea born in 3. Jeremy born in 4. Craig born in 5. Elizabeth born in" }
    \label{logitLensSample}
    \vspace{0.5em}
    \begin{tabular}{c|c}
        \toprule
        MLP & Top-3 Tokens \\
        \midrule
        0 & order, the, particular \\
        1 & the, order, a \\
        2 & the, order, a \\
        3 & the, order, accordance \\
        4 & order, the, front \\
        5 & 18, 3, 2 \\
        6 & 5, 3, 2 \\
        7 & 3, 5, 2 \\
        8 & 5, 6, 4 \\
        9 & 6, 5, 7 \\
        10 & 6, 7, 8 \\
        11 & 6, 7, 1 \\
        \bottomrule
    \end{tabular}
\end{table}

\begin{table}
    \centering
    \caption{Logit Lens- "8 9 10 11" }
    \label{logitLensSample_pureNumerals}
    \vspace{0.5em}
    \begin{tabular}{c|c}
        \toprule
        MLP & Top-3 Tokens \\
        \midrule
        0 & th, 11, 11 \\
        1 & th, 11, 45 \\
        2 & th, 30, 45 \\
        3 & 30, 45, 34 \\
        4 & 45, 34, th \\
        5 & votes, ., 9 \\
        6 & 9, ., 11 \\
        7 & 11, 9, 1 \\
        8 & 11, 12, 111 \\
        9 & 12, 11, 12 \\
        10 & 12, 13, 12 \\
        11 & 12, 13, \textbackslash n \\
        \bottomrule
    \end{tabular}
\end{table}

\begin{table}
    \centering
    \caption{Logit Lens- "seven eight nine ten" }
    \label{logitLens_nw}
    \vspace{0.5em}
    \begin{tabular}{c|c}
        \toprule
        MLP & Top-3 Tokens \\
        \midrule
        0 & thousand, ten, years \\
        1 & thousand, fold, hundred \\
        2 & thousand, percent, minutes \\
        3 & thousand, percent, years \\
        4 & thousand, percent, million \\
        5 & thousand, ths, million \\
        6 & thousand, million, years \\
        7 & thousand, ths, 9 \\
        8 & nine, 11, 9 \\
        9 & eleven, 11, twelve \\
        10 & eleven, 11, twelve \\
        11 & eleven, twelve, 11 \\
        \bottomrule
    \end{tabular}
\end{table}

\begin{table}
    \centering
    \caption{Logit Lens- "May June July August" }
    \label{logitLens_months}
    \vspace{0.5em}
    \begin{tabular}{c|c}
        \toprule
        MLP & Top-3 Tokens \\
        \midrule
        0 & Aug, August, 2017 \\
        1 & 2017, 2014, Aug \\
        2 & 2014, 2017, 2015 \\
        3 & 2017, 2014, 2015 \\
        4 & 2014, 2017, 2018 \\
        5 & 2014, 2013, 2018 \\
        6 & September, December, August \\
        7 & December, September, August \\
        8 & August, September, October \\
        9 & September, August, October \\
        10 & September, August, October \\
        11 & September, August, October \\
        \bottomrule
    \end{tabular}
\end{table}




\section{Attention Pattern Extended Results for GPT-2 Small}
\label{appendix:imptHeads}

When we run attention pattern analysis on sequences comprised solely of sequence member tokens such as “1 2 3 4”, there are no other ‘non-sequence member’ words to compare to, so it is hard to tell what ‘type’ of token each head is attending to. Thus, we measure what types of tokens the heads attend to by using prompts that contained these sequences within other types of tokens, such as "Table lost in March. Lamp lost in April."

\begin{figure*}[ht]  
  \centering
 \includegraphics[scale=0.4]{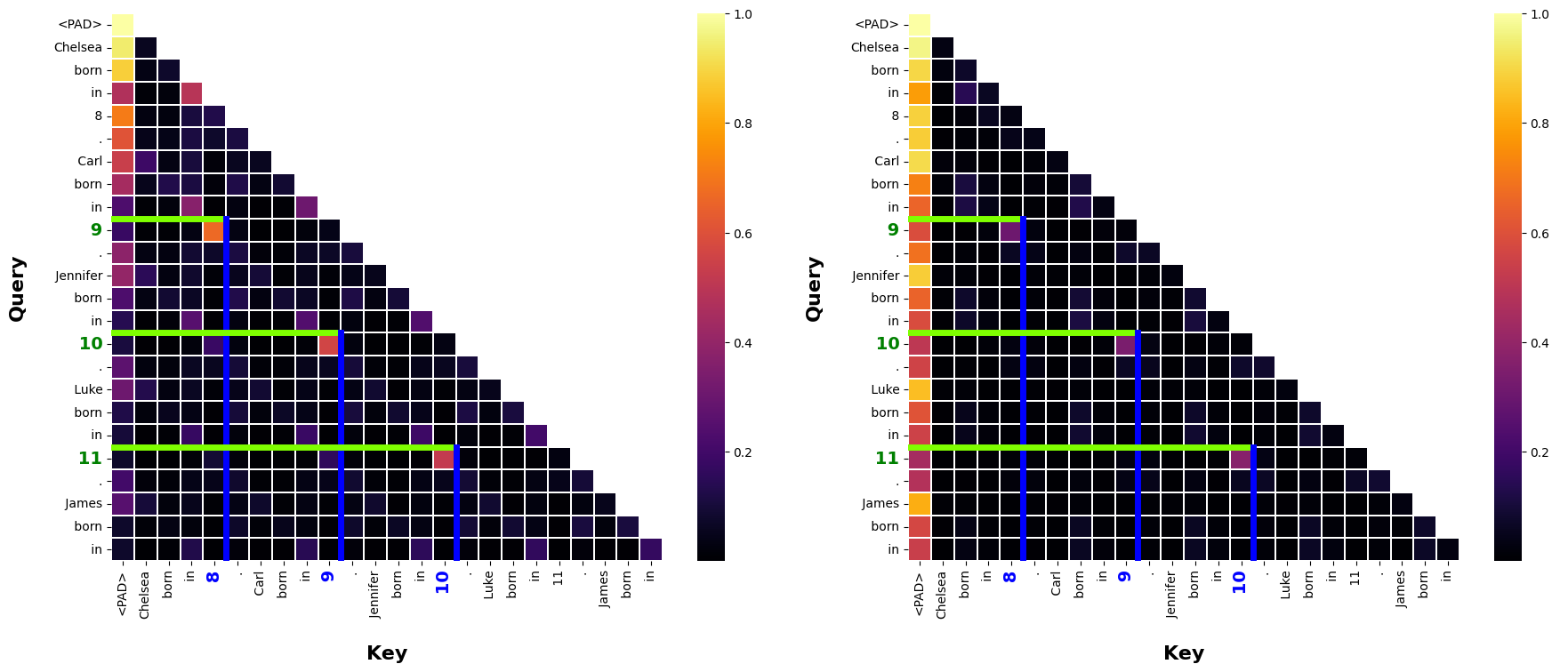}
  \caption{Attention Patterns for Numerals of (a) Head 1.5 and (b) Head 4.4. Lighter colors mean higher attention values. For each of these detection patterns, the \textcolor{darkgreen}{query is shown in green}, and the \textcolor{blue}{key is shown in blue}. We observe that numerals attend to numerals, but they are not considered general "similarity detection heads" as non-number token types do not attend to their similar or same token types. }
  \label{fig:attn_early_numerals_oneline}
\end{figure*}

\begin{figure*}[ht]
  \centering
 \includegraphics[scale=0.4]{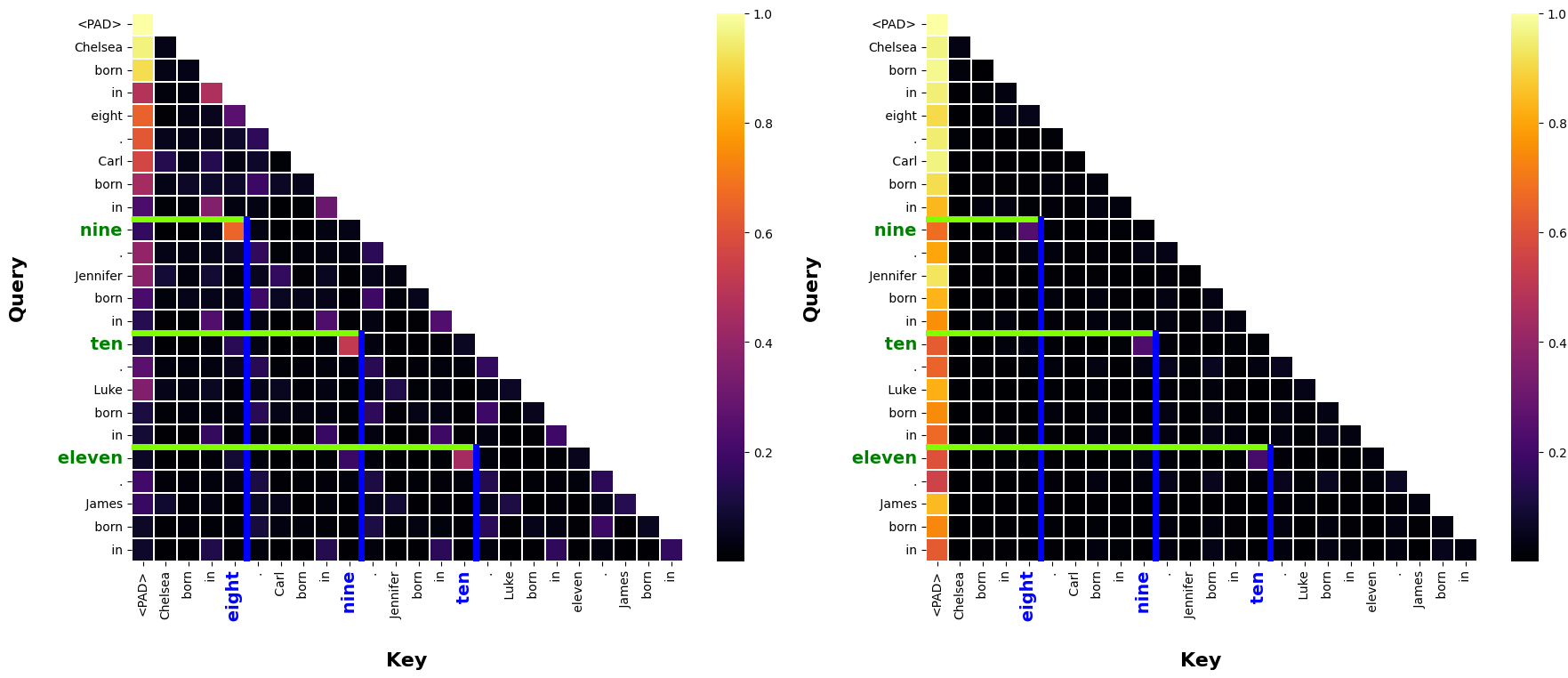}
  \caption{Attention Patterns for Number Words of (a) Head 1.5 and (b) Head 4.4. We observe that number words attend to number words (for head 4.4, these scores are in \textcolor{darkblue}{dark blue}). We also observe that the attention scores here are less than they are for digits, suggesting that head 4.4 is more important for digit detection, which is consistent with its importance for the digits task over the number words task as shown in Table \ref{circHeadsDrop}.  }
  \label{fig:attn_early_numwords_oneline}
\end{figure*}

\begin{figure*}[ht]
  \centering
 \includegraphics[scale=0.2]{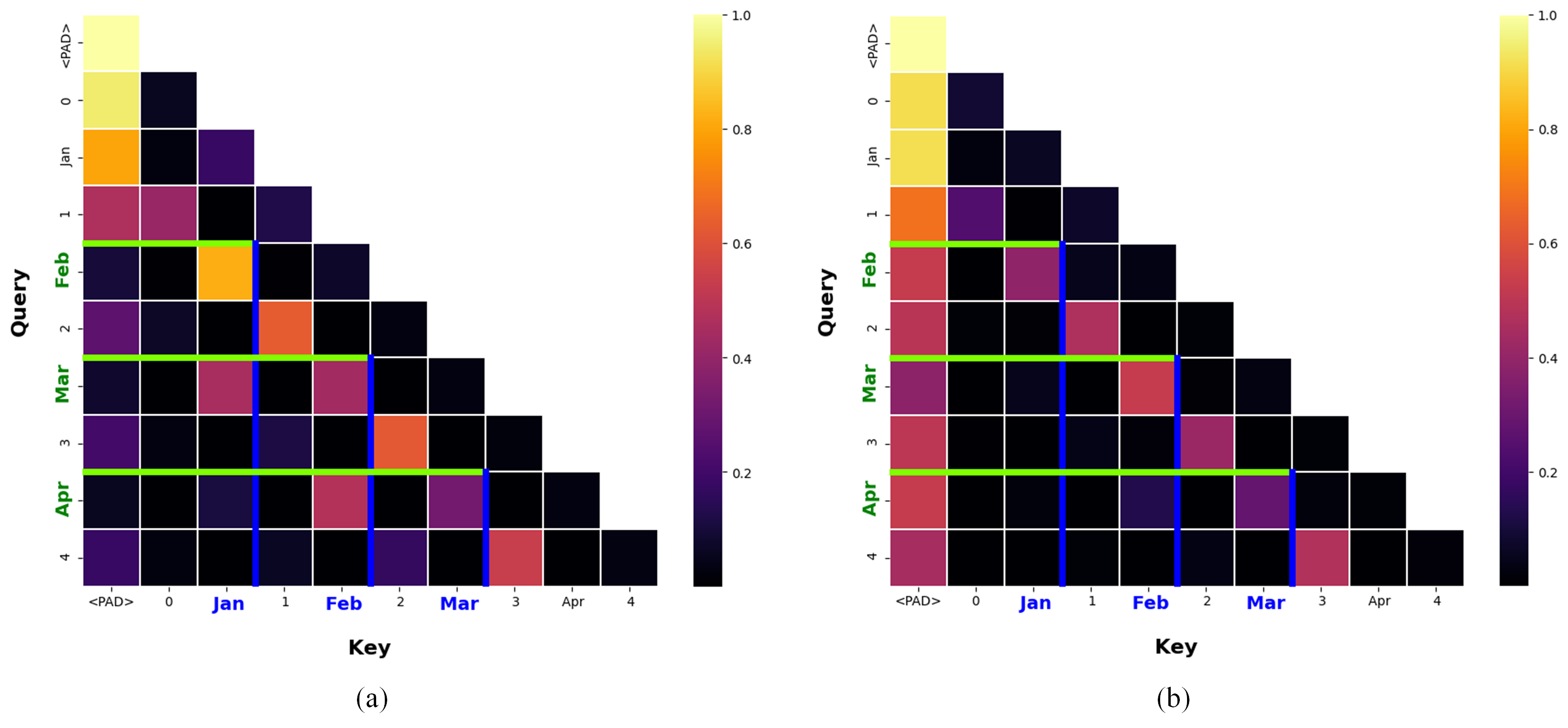}
  \caption{Attention Patterns for (a) Head 1.5 and (b) Head 4.4. We observe that digits attend to digits, and that months attend to months. In general, they appear to be adjacent sequence member detection heads. }
  \label{fig:attnpat_early_mixed_digits_months}
\end{figure*}

\textbf{Sequence Member Detection Heads Details. } We discovered a "similar member" detection head, 1.5, and a "sequence member detection", 4.4, both shown in Figure \ref{fig:attn_early_numerals_oneline}, where numerals attend to previous numerals, and in Figure \ref{fig:attn_early_numwords_oneline}, where number words attend to previous number words. 
To discern whether these heads are "similarity detection" heads in general, or are more specific to detecting sequence members such as numbers, these figures show
that not all token types attend to their similar types; for instance, names do not attend to names. We also do not observe every token attending to a previous position k tokens back (where k is an integer), so we do not conclude that these heads also act as previous token heads.
Additionally, Figure \ref{fig:attnpat_early_mixed_digits_months} shows that when both Numerals and Months are in sequence order, the heads attend to both Numerals and Months. 


\section{Broader Impact Statement}
This paper analyzes shared computational structures in a transformer language model for semantically similar sequence continuation tasks. The goal is to advance the mechanistic understanding of how these models represent and process concepts.

While increasing interpretability can lead to positive outcomes like safer AI systems, there is also potential for dual use if mechanisms are uncovered that could be exploited. For example, the knowledge that models share components between analogous tasks could perhaps be misused to more effectively generate synthetic harmful text by exploiting knowledge about one task to another when performing model editing.
To mitigate these risks, the code, data and findings from this work will be published under an open source license. However, malicious actors may not respect licensing limits, so additional precautions are warranted. The authors recommend coordinated disclosure and further research into techniques like model editing that could leverage these insights to align language models, making them more robust and less susceptible to misuse.

More broadly, the interpretability community should continue working to establish norms and best practices around responsible disclosure of model internals. Understanding the inner workings of systems like LLMs is not an end in itself, but rather a means towards developing language technology that is trustworthy, ethical, and benefits society. Researchers in this space should thoughtfully consider broader impacts as core to their work.

%% file: main.bbl
\begin{thebibliography}{42}
\expandafter\ifx\csname natexlab\endcsname\relax\def\natexlab#1{#1}\fi

\bibitem[{Amodei et~al.(2016)Amodei, Olah, Steinhardt, Christiano, Schulman,
  and Man{\'e}}]{Amodei2016ConcretePI}
Dario Amodei, Chris Olah, Jacob Steinhardt, Paul Christiano, John Schulman, and
  Dan Man{\'e}. 2016.
\newblock Concrete problems in ai safety.
\newblock \emph{arXiv: Learning}, abs/1606.06565.

\bibitem[{Barez et~al.(2023)Barez, Hasanbieg, and Abbate}]{barez2023iii}
Fazl Barez, Hosien Hasanbieg, and Alesandro Abbate. 2023.
\newblock \href {http://arxiv.org/abs/2304.11593} {System iii: Learning with
  domain knowledge for safety constraints}.

\bibitem[{Barredo~Arrieta et~al.(2020)Barredo~Arrieta, Díaz-Rodríguez,
  Del~Ser, Bennetot, Tabik, Barbado, Garcia, Gil-Lopez, Molina, Benjamins,
  Chatila, and Herrera}]{BarredoArrieta2020ExplainableAI}
Alejandro Barredo~Arrieta, Natalia Díaz-Rodríguez, Javier Del~Ser, Adrien
  Bennetot, Siham Tabik, Alberto Barbado, Salvador Garcia, Sergio Gil-Lopez,
  Daniel Molina, Richard Benjamins, Raja Chatila, and Francisco Herrera. 2020.
\newblock \href {https://doi.org/10.1016/j.inffus.2019.12.012} {Explainable
  artificial intelligence (xai): Concepts, taxonomies, opportunities and
  challenges toward responsible ai}.
\newblock \emph{Information Fusion}, 58:82--115.

\bibitem[{Brown et~al.(2020)Brown, Mann, Ryder, Subbiah, Kaplan, Dhariwal,
  Neelakantan, Shyam, Sastry, Askell et~al.}]{brown2020language}
Tom Brown, Benjamin Mann, Nick Ryder, Melanie Subbiah, Jared~D Kaplan, Prafulla
  Dhariwal, Arvind Neelakantan, Pranav Shyam, Girish Sastry, Amanda Askell,
  et~al. 2020.
\newblock Language models are few-shot learners.
\newblock \emph{Advances in neural information processing systems},
  33:1877--1901.

\bibitem[{Bubeck et~al.(2023)Bubeck, Chandrasekaran, Eldan, Gehrke, Horvitz,
  Kamar, Lee, Lee, Li, Lundberg, Nori, Palangi, Ribeiro, and
  Zhang}]{bubeck2023sparks}
Sébastien Bubeck, Varun Chandrasekaran, Ronen Eldan, Johannes Gehrke, Eric
  Horvitz, Ece Kamar, Peter Lee, Yin~Tat Lee, Yuanzhi Li, Scott Lundberg,
  Harsha Nori, Hamid Palangi, Marco~Tulio Ribeiro, and Yi~Zhang. 2023.
\newblock \href {http://arxiv.org/abs/2303.12712} {Sparks of artificial general
  intelligence: Early experiments with gpt-4}.

\bibitem[{Caldarini et~al.(2022)Caldarini, Jaf, and
  McGarry}]{caldarini2022literature}
Guendalina Caldarini, Sardar Jaf, and Kenneth McGarry. 2022.
\newblock A literature survey of recent advances in chatbots.
\newblock \emph{Information}, 13(1):41.

\bibitem[{Conmy et~al.(2023)Conmy, Mavor-Parker, Lynch, Heimersheim, and
  Garriga-Alonso}]{conmy2023towards}
Arthur Conmy, Augustine~N Mavor-Parker, Aengus Lynch, Stefan Heimersheim, and
  Adri{\`a} Garriga-Alonso. 2023.
\newblock Towards automated circuit discovery for mechanistic interpretability.
\newblock \emph{arXiv preprint arXiv:2304.14997}.

\bibitem[{Dar et~al.(2022)Dar, Geva, Gupta, and Berant}]{dar2022analyzing}
Guy Dar, Mor Geva, Ankit Gupta, and Jonathan Berant. 2022.
\newblock \href {http://arxiv.org/abs/2209.02535} {Analyzing transformers in
  embedding space}.

\bibitem[{Elhage et~al.(2022)Elhage, Hume, Olsson, Schiefer, Henighan, Kravec,
  Hatfield-Dodds, Lasenby, Drain, Chen, Grosse, McCandlish, Kaplan, Amodei,
  Wattenberg, and Olah}]{elhage2022superposition}
Nelson Elhage, Tristan Hume, Catherine Olsson, Nicholas Schiefer, Tom Henighan,
  Shauna Kravec, Zac Hatfield-Dodds, Robert Lasenby, Dawn Drain, Carol Chen,
  Roger Grosse, Sam McCandlish, Jared Kaplan, Dario Amodei, Martin Wattenberg,
  and Christopher Olah. 2022.
\newblock Toy models of superposition.
\newblock \emph{Transformer Circuits Thread}.
\newblock Https://transformer-circuits.pub/2022/toy\_model/index.html.

\bibitem[{Elhage et~al.(2021)Elhage, Nanda, Olsson, Henighan, Joseph, Mann,
  Askell, Bai, Chen, Conerly, DasSarma, Drain, Ganguli, Hatfield-Dodds,
  Hernandez, Jones, Kernion, Lovitt, Ndousse, Amodei, Brown, Clark, Kaplan,
  McCandlish, and Olah}]{elhage2021mathematical}
Nelson Elhage, Neel Nanda, Catherine Olsson, Tom Henighan, Nicholas Joseph, Ben
  Mann, Amanda Askell, Yuntao Bai, Anna Chen, Tom Conerly, Nova DasSarma, Dawn
  Drain, Deep Ganguli, Zac Hatfield-Dodds, Danny Hernandez, Andy Jones, Jackson
  Kernion, Liane Lovitt, Kamal Ndousse, Dario Amodei, Tom Brown, Jack Clark,
  Jared Kaplan, Sam McCandlish, and Chris Olah. 2021.
\newblock A mathematical framework for transformer circuits.
\newblock \emph{Transformer Circuits Thread}.
\newblock Https://transformer-circuits.pub/2021/framework/index.html.

\bibitem[{Engels et~al.(2024)Engels, Liao, Michaud, Gurnee, and
  Tegmark}]{engels2024language}
Joshua Engels, Isaac Liao, Eric~J. Michaud, Wes Gurnee, and Max Tegmark. 2024.
\newblock \href {http://arxiv.org/abs/2405.14860} {Not all language model
  features are linear}.

\bibitem[{Foote et~al.(2023)Foote, Nanda, Kran, Konstas, Cohen, and
  Barez}]{foote2023neuron}
Alex Foote, Neel Nanda, Esben Kran, Ioannis Konstas, Shay Cohen, and Fazl
  Barez. 2023.
\newblock \href {http://arxiv.org/abs/2305.19911} {Neuron to graph:
  Interpreting language model neurons at scale}.
\newblock In \emph{Proceedings of the Trustworthy and Reliable Large-Scale
  Machine Learning Models Workshop at ICLR}.

\bibitem[{Geva et~al.(2020)Geva, Schuster, Berant, and
  Levy}]{geva2020transformer}
Mor Geva, Roei Schuster, Jonathan Berant, and Omer Levy. 2020.
\newblock Transformer feed-forward layers are key-value memories.
\newblock \emph{arXiv preprint arXiv:2012.14913}.

\bibitem[{Goldowsky-Dill et~al.(2023)Goldowsky-Dill, MacLeod, Sato, and
  Arora}]{goldowskydill2023localizing}
Nicholas Goldowsky-Dill, Chris MacLeod, Lucas Sato, and Aryaman Arora. 2023.
\newblock \href {http://arxiv.org/abs/2304.05969} {Localizing model behavior
  with path patching}.

\bibitem[{Gould et~al.(2023)Gould, Ong, Ogden, and Conmy}]{gould2023successor}
Rhys Gould, Euan Ong, George Ogden, and Arthur Conmy. 2023.
\newblock \href {http://arxiv.org/abs/2312.09230} {Successor heads: Recurring,
  interpretable attention heads in the wild}.

\bibitem[{Gurnee and Tegmark(2023)}]{gurnee2023language}
Wes Gurnee and Max Tegmark. 2023.
\newblock \href {http://arxiv.org/abs/2310.02207} {Language models represent
  space and time}.

\bibitem[{Hanna et~al.(2023)Hanna, Liu, and Variengien}]{hanna2023does}
Michael Hanna, Ollie Liu, and Alexandre Variengien. 2023.
\newblock \href {http://arxiv.org/abs/2305.00586} {How does gpt-2 compute
  greater-than?: Interpreting mathematical abilities in a pre-trained language
  model}.

\bibitem[{Hendrycks and Mazeika(2022)}]{hendrycks2022x}
Dan Hendrycks and Mantas Mazeika. 2022.
\newblock X-risk analysis for ai research.
\newblock \emph{arXiv preprint arXiv:2206.05862}.

\bibitem[{Hoelscher-Obermaier et~al.(2023)Hoelscher-Obermaier, Persson, Kran,
  Konstas, and Barez}]{hoelscherobermaier2023detecting}
Jason Hoelscher-Obermaier, Julia Persson, Esben Kran, Ioannis Konstas, and Fazl
  Barez. 2023.
\newblock \href {http://arxiv.org/abs/2305.17553} {Detecting edit failures in
  large language models: An improved specificity benchmark}.

\bibitem[{Huh et~al.(2024)Huh, Cheung, Wang, and Isola}]{huh2024platonic}
Minyoung Huh, Brian Cheung, Tongzhou Wang, and Phillip Isola. 2024.
\newblock \href {http://arxiv.org/abs/2405.07987} {The platonic representation
  hypothesis}.

\bibitem[{Kojima et~al.(2024)Kojima, Okimura, Iwasawa, Yanaka, and
  Matsuo}]{kojima2024multilingual}
Takeshi Kojima, Itsuki Okimura, Yusuke Iwasawa, Hitomi Yanaka, and Yutaka
  Matsuo. 2024.
\newblock \href {http://arxiv.org/abs/2404.02431} {On the multilingual ability
  of decoder-based pre-trained language models: Finding and controlling
  language-specific neurons}.

\bibitem[{Marks et~al.(2023)Marks, Abdullah, Mendez, Arike, Torr, and
  Barez}]{marks2023interpreting}
Luke Marks, Amir Abdullah, Luna Mendez, Rauno Arike, Philip Torr, and Fazl
  Barez. 2023.
\newblock \href {http://arxiv.org/abs/2310.08164} {Interpreting reward models
  in rlhf-tuned language models using sparse autoencoders}.

\bibitem[{Meng et~al.(2023)Meng, Bau, Andonian, and
  Belinkov}]{meng2023locating}
Kevin Meng, David Bau, Alex Andonian, and Yonatan Belinkov. 2023.
\newblock \href {http://arxiv.org/abs/2202.05262} {Locating and editing factual
  associations in gpt}.

\bibitem[{Merullo et~al.(2023)Merullo, Eickhoff, and
  Pavlick}]{merullo2023circuit}
Jack Merullo, Carsten Eickhoff, and Ellie Pavlick. 2023.
\newblock \href {http://arxiv.org/abs/2310.08744} {Circuit component reuse
  across tasks in transformer language models}.

\bibitem[{Miceli-Barone et~al.(2023)Miceli-Barone, Barez, Konstas, and
  Cohen}]{micelibarone2023larger}
Antonio~Valerio Miceli-Barone, Fazl Barez, Ioannis Konstas, and Shay~B. Cohen.
  2023.
\newblock \href {http://arxiv.org/abs/2305.15507} {The larger they are, the
  harder they fail: Language models do not recognize identifier swaps in
  python}.

\bibitem[{Mu and Andreas(2020)}]{mu2020compositional}
Jesse Mu and Jacob Andreas. 2020.
\newblock Compositional explanations of neurons.
\newblock \emph{Advances in Neural Information Processing Systems},
  33:17153--17163.

\bibitem[{Nanda et~al.(2023)Nanda, Chan, Lieberum, Smith, and
  Steinhardt}]{nanda2023progress}
Neel Nanda, Lawrence Chan, Tom Lieberum, Jess Smith, and Jacob Steinhardt.
  2023.
\newblock \href {http://arxiv.org/abs/2301.05217} {Progress measures for
  grokking via mechanistic interpretability}.

\bibitem[{Nostalgebraist(2020)}]{nostalgebraist2020interpreting}
Nostalgebraist. 2020.
\newblock Interpreting gpt: The logit lens.
\newblock
  \url{https://www.alignmentforum.org/posts/AcKRB8wDpdaN6v6ru/interpreting-gpt-the-logit-lens}.
\newblock Accessed: 14 December 2023.

\bibitem[{Olsson et~al.(2022)Olsson, Elhage, Nanda, Joseph, DasSarma, Henighan,
  Mann, Askell, Bai, Chen, Conerly, Drain, Ganguli, Hatfield-Dodds, Hernandez,
  Johnston, Jones, Kernion, Lovitt, Ndousse, Amodei, Brown, Clark, Kaplan,
  McCandlish, and Olah}]{olsson2022context}
Catherine Olsson, Nelson Elhage, Neel Nanda, Nicholas Joseph, Nova DasSarma,
  Tom Henighan, Ben Mann, Amanda Askell, Yuntao Bai, Anna Chen, Tom Conerly,
  Dawn Drain, Deep Ganguli, Zac Hatfield-Dodds, Danny Hernandez, Scott
  Johnston, Andy Jones, Jackson Kernion, Liane Lovitt, Kamal Ndousse, Dario
  Amodei, Tom Brown, Jack Clark, Jared Kaplan, Sam McCandlish, and Chris Olah.
  2022.
\newblock In-context learning and induction heads.
\newblock \emph{Transformer Circuits Thread}.
\newblock
  Https://transformer-circuits.pub/2022/in-context-learning-and-induction-heads/index.html.

\bibitem[{Park et~al.(2024)Park, Choe, Jiang, and Veitch}]{park2024geometry}
Kiho Park, Yo~Joong Choe, Yibo Jiang, and Victor Veitch. 2024.
\newblock \href {http://arxiv.org/abs/2406.01506} {The geometry of categorical
  and hierarchical concepts in large language models}.

\bibitem[{Quirke and Barez(2023)}]{quirke2023understanding}
Philip Quirke and Fazl Barez. 2023.
\newblock \href {http://arxiv.org/abs/2310.13121} {Understanding addition in
  transformers}.

\bibitem[{Radford et~al.(2019)Radford, Wu, Child, Luan, Amodei, and
  Sutskever}]{radford_language_2019}
Alec Radford, Jeff Wu, Rewon Child, D.~Luan, Dario Amodei, and Ilya Sutskever.
  2019.
\newblock Language models are unsupervised multitask learners.

\bibitem[{Stolfo et~al.(2023)Stolfo, Belinkov, and
  Sachan}]{stolfo-etal-2023-mechanistic}
Alessandro Stolfo, Yonatan Belinkov, and Mrinmaya Sachan. 2023.
\newblock \href {https://doi.org/10.18653/v1/2023.emnlp-main.435} {A
  mechanistic interpretation of arithmetic reasoning in language models using
  causal mediation analysis}.
\newblock In \emph{Proceedings of the 2023 Conference on Empirical Methods in
  Natural Language Processing}, pages 7035--7052, Singapore. Association for
  Computational Linguistics.

\bibitem[{Templeton et~al.(2024)Templeton, Conerly, Marcus, Lindsey, Bricken,
  Chen, Pearce, Citro, Ameisen, Jones, Cunningham, Turner, McDougall,
  MacDiarmid, Freeman, Sumers, Rees, Batson, Jermyn, Carter, Olah, and
  Henighan}]{templeton2024scaling}
Adly Templeton, Tom Conerly, Jonathan Marcus, Jack Lindsey, Trenton Bricken,
  Brian Chen, Adam Pearce, Craig Citro, Emmanuel Ameisen, Andy Jones, Hoagy
  Cunningham, Nicholas~L Turner, Callum McDougall, Monte MacDiarmid, C.~Daniel
  Freeman, Theodore~R. Sumers, Edward Rees, Joshua Batson, Adam Jermyn, Shan
  Carter, Chris Olah, and Tom Henighan. 2024.
\newblock \href
  {https://transformer-circuits.pub/2024/scaling-monosemanticity/index.html}
  {Scaling monosemanticity: Extracting interpretable features from claude 3
  sonnet}.
\newblock \emph{Transformer Circuits Thread}.

\bibitem[{Touvron et~al.(2023)Touvron, Martin, Stone, Albert, Almahairi,
  Babaei, Bashlykov, Batra, Bhargava, Bhosale, Bikel, Blecher, Ferrer, Chen,
  Cucurull, Esiobu, Fernandes, Fu, Fu, Fuller, Gao, Goswami, Goyal, Hartshorn,
  Hosseini, Hou, Inan, Kardas, Kerkez, Khabsa, Kloumann, Korenev, Koura,
  Lachaux, Lavril, Lee, Liskovich, Lu, Mao, Martinet, Mihaylov, Mishra,
  Molybog, Nie, Poulton, Reizenstein, Rungta, Saladi, Schelten, Silva, Smith,
  Subramanian, Tan, Tang, Taylor, Williams, Kuan, Xu, Yan, Zarov, Zhang, Fan,
  Kambadur, Narang, Rodriguez, Stojnic, Edunov, and
  Scialom}]{touvron2023llama2openfoundation}
Hugo Touvron, Louis Martin, Kevin Stone, Peter Albert, Amjad Almahairi, Yasmine
  Babaei, Nikolay Bashlykov, Soumya Batra, Prajjwal Bhargava, Shruti Bhosale,
  Dan Bikel, Lukas Blecher, Cristian~Canton Ferrer, Moya Chen, Guillem
  Cucurull, David Esiobu, Jude Fernandes, Jeremy Fu, Wenyin Fu, Brian Fuller,
  Cynthia Gao, Vedanuj Goswami, Naman Goyal, Anthony Hartshorn, Saghar
  Hosseini, Rui Hou, Hakan Inan, Marcin Kardas, Viktor Kerkez, Madian Khabsa,
  Isabel Kloumann, Artem Korenev, Punit~Singh Koura, Marie-Anne Lachaux,
  Thibaut Lavril, Jenya Lee, Diana Liskovich, Yinghai Lu, Yuning Mao, Xavier
  Martinet, Todor Mihaylov, Pushkar Mishra, Igor Molybog, Yixin Nie, Andrew
  Poulton, Jeremy Reizenstein, Rashi Rungta, Kalyan Saladi, Alan Schelten, Ruan
  Silva, Eric~Michael Smith, Ranjan Subramanian, Xiaoqing~Ellen Tan, Binh Tang,
  Ross Taylor, Adina Williams, Jian~Xiang Kuan, Puxin Xu, Zheng Yan, Iliyan
  Zarov, Yuchen Zhang, Angela Fan, Melanie Kambadur, Sharan Narang, Aurelien
  Rodriguez, Robert Stojnic, Sergey Edunov, and Thomas Scialom. 2023.
\newblock \href {http://arxiv.org/abs/2307.09288} {Llama 2: Open foundation and
  fine-tuned chat models}.

\bibitem[{Turner et~al.(2023)Turner, Thiergart, Udell, Leech, Mini, and
  MacDiarmid}]{turner2023activation}
Alexander~Matt Turner, Lisa Thiergart, David Udell, Gavin Leech, Ulisse Mini,
  and Monte MacDiarmid. 2023.
\newblock \href {http://arxiv.org/abs/2308.10248} {Activation addition:
  Steering language models without optimization}.

\bibitem[{Vaswani et~al.(2017)Vaswani, Shazeer, Parmar, Uszkoreit, Jones,
  Gomez, Kaiser, and Polosukhin}]{vaswani2017attention}
Ashish Vaswani, Noam Shazeer, Niki Parmar, Jakob Uszkoreit, Llion Jones,
  Aidan~N Gomez, {\L}ukasz Kaiser, and Illia Polosukhin. 2017.
\newblock Attention is all you need.
\newblock \emph{Advances in neural information processing systems}, 30.

\bibitem[{Vig et~al.(2020)Vig, Gehrmann, Belinkov, Qian, Nevo, Singer, and
  Shieber}]{vig2020investigating}
Jesse Vig, Sebastian Gehrmann, Yonatan Belinkov, Sharon Qian, Daniel Nevo,
  Yaron Singer, and Stuart Shieber. 2020.
\newblock Investigating gender bias in language models using causal mediation
  analysis.
\newblock \emph{Advances in neural information processing systems},
  33:12388--12401.

\bibitem[{Wang et~al.(2022)Wang, Variengien, Conmy, Shlegeris, and
  Steinhardt}]{wang2022interpretability}
Kevin Wang, Alexandre Variengien, Arthur Conmy, Buck Shlegeris, and Jacob
  Steinhardt. 2022.
\newblock \href {http://arxiv.org/abs/2211.00593} {Interpretability in the
  wild: a circuit for indirect object identification in gpt-2 small}.

\bibitem[{Wendler et~al.(2024)Wendler, Veselovsky, Monea, and
  West}]{wendler2024llamas}
Chris Wendler, Veniamin Veselovsky, Giovanni Monea, and Robert West. 2024.
\newblock \href {http://arxiv.org/abs/2402.10588} {Do llamas work in english?
  on the latent language of multilingual transformers}.

\bibitem[{Zhang et~al.(2022)Zhang, Xing, Zou et~al.}]{Zhang2022ShiftingML}
Audrey Zhang, Liang Xing, James Zou, et~al. 2022.
\newblock \href {https://doi.org/10.1038/s41551-022-00898-y} {Shifting machine
  learning for healthcare from development to deployment and from models to
  data}.
\newblock \emph{Nature Biomedical Engineering}, 6:1330--1345.

\bibitem[{Zhuang et~al.(2020)Zhuang, Qi, Duan, Xi, Zhu, Zhu, Xiong, and
  He}]{zhuang2020comprehensive}
Fuzhen Zhuang, Zhiyuan Qi, Keyu Duan, Dongbo Xi, Yongchun Zhu, Hengshu Zhu, Hui
  Xiong, and Qing He. 2020.
\newblock \href {http://arxiv.org/abs/1911.02685} {A comprehensive survey on
  transfer learning}.

\end{thebibliography}
